\documentclass{article} 
\usepackage[preprint]{colm2026_conference}
\usepackage{wrapfig}
\usepackage{microtype}
\usepackage{hyperref}
\usepackage{url}
\usepackage{booktabs}
\usepackage{graphicx}
\usepackage{amsmath}
\usepackage[table,dvipsnames]{xcolor}
\usepackage{colortbl}
\usepackage{subcaption}
\definecolor{lightgray}{gray}{0.95}
\usepackage{tablefootnote}
\usepackage{textcomp}
\usepackage{tabularx}
\usepackage{pifont}
\usepackage{makecell}
\usepackage{multicol}
\usepackage{multirow}
\usepackage{musicography}
\usepackage[most]{tcolorbox}
\usepackage{enumitem}
\usepackage{algorithm}
\usepackage{algorithmic}
\lstdefinestyle{prompt}{
    basicstyle=\ttfamily\fontsize{7pt}{8pt}\selectfont,
    frame=none,
    breaklines=true,
    backgroundcolor=\color{lightgray},
    breakatwhitespace=true,
    breakindent=0pt,
    escapeinside={(*@}{@*)},
    numbers=none,
    numbersep=5pt,
    xleftmargin=5pt,
}
\tcbset{
  aibox/.style={
    top=10pt,
    colback=white,
    colframe=black,
    colbacktitle=black,
    enhanced,
    center,
    attach boxed title to top left={yshift=-0.1in,xshift=0.15in},
    boxed title style={boxrule=0pt,colframe=white,},
  }
}
\newtcolorbox{AIbox}[2][]{aibox, title=#2,#1}

\usepackage{lineno}

\newcommand{\weihuanote}[1]{}
\newcommand{\METHOD}{AdaExplore}

\definecolor{darkblue}{rgb}{0, 0, 0.5}
\hypersetup{colorlinks=true, citecolor=darkblue, linkcolor=darkblue, urlcolor=darkblue}

\title{\METHOD: Failure-Driven Adaptation and Diversity-Preserving Search for Efficient Kernel Generation}

\author{Weihua Du$^1$, Jingming Zhuo$^2$, Yixin Dong$^1$, Andre He$^1$, \\
\textbf{Weiwei Sun$^1$, Zeyu Zheng$^1$, Manupa Karunaratne$^3$, Ivan Fox$^3$,} \\
\textbf{Tim Dettmers$^1$, Tianqi Chen$^1$, Yiming Yang$^1$, Sean Welleck$^1$} \\
$^1$Carnegie Mellon University, $^2$University of Washington, $^3$Arm Ltd.\\
\texttt{\{weihuad, swelleck\}@cs.cmu.edu}
}

\begin{document}

\ifcolmsubmission
\linenumbers
\fi

\maketitle
\begin{abstract}
Recent large language model (LLM) agents have shown promise in using execution feedback for test-time adaptation. However, robust self-improvement remains far from solved: most approaches still treat each problem instance independently, without accumulating reusable knowledge.
This limitation is particularly pronounced in domain-specific languages such as Triton, which are underrepresented in LLM pretraining data. Their strict constraints and non-linear optimization landscape further make naive generation and local refinement unreliable.
We propose \textbf{\METHOD}, an agent framework that enables \textbf{self-improvement via accumulated execution feedback} for performance-critical kernel code generation through two complementary stages: \textbf{failure-driven adaptation} and \textbf{diversity-preserving search}, jointly improving correctness and optimization performance without additional fine-tuning or external knowledge.
In the adaptation stage, the agent synthesizes tasks and converts recurring failures into a reusable memory of validity rules, helping subsequent generations remain within the feasible set.
In the search stage, the agent organizes candidate kernels as a tree and alternates between small local refinements and larger structural regeneration, allowing it to explore the optimization landscape beyond local optima.
Experiments on kernel runtime optimization benchmarks validate these gains: \METHOD{} achieves 3.11$\times$ and 1.62$\times$ speedups on KernelBench Level-2 and Level-3, respectively, within 100 steps and using GPT-5-mini, and continues to improve with additional computation.
\ifcolmsubmission\else
Our implementation is publicly available at \href{https://github.com/StigLidu/AdaExplore}{https://github.com/StigLidu/AdaExplore}.
\fi
\end{abstract}

\section{Introduction}

Large language models (LLMs) have rapidly evolved into capable coding agents, achieving strong performance on tasks such as bug fixing, refactoring, and unit testing \citep{chen2021evaluating, li2022competition, nijkamp2022codegen}. Recent work further extends LLMs to tool-augmented agents that iteratively synthesize and debug programs \citep{wang2025agents, qian2024chatdev, zhang2024codeagent}.
In this work, we study \textit{code runtime optimization for GPU kernels} in low-level programming frameworks such as Triton~\citep{ouyang2025kernelbench, li2025tritonbench}. Unlike conventional code generation, which focuses on functional correctness, code optimization requires satisfying correctness as a hard constraint while optimizing runtime performance. This explicit performance objective provides a natural feedback signal for sustained improvement.

\begin{figure*}[t]
    \centering
    \vspace{-8mm}
    \includegraphics[width=1\linewidth]{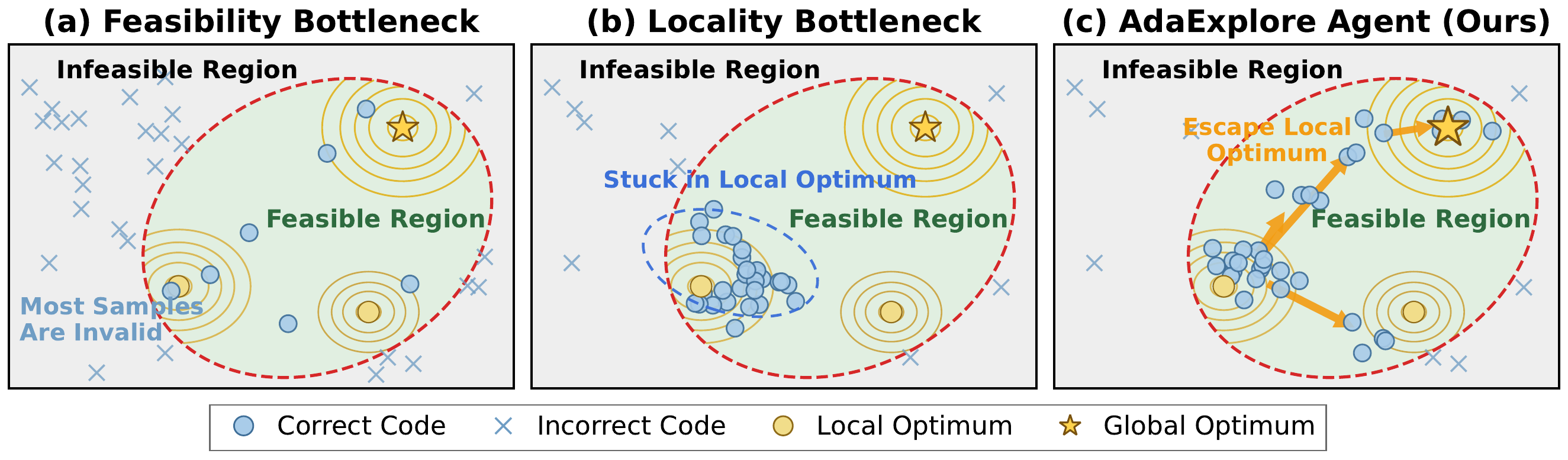}
    \vspace{-4mm}
    \caption{\textbf{Illustration of Kernel Optimization Bottlenecks.} \textbf{(a)} Most generated kernels are invalid due to limited training coverage; \textbf{(b)} Kernel refinement may be stuck in local optima; \textbf{(c)} Our agent, \textbf{\METHOD{}}, can learn skills from failures to prevent pitfalls, and apply diversity-preserving search for global optima exploration.}
    \label{fig:insight}
    \vspace{-3mm}
\end{figure*}

However, despite this favorable signal, the search space remains highly challenging. First, as illustrated in Figure~\ref{fig:insight}a, the \textbf{feasibility boundary is sharp}: small errors in syntax, memory access, or parallelization often lead to compilation failures or runtime errors. This challenge is further exacerbated by the limited availability of training data for low-level programming languages, which weakens the model's prior over valid implementations and results in a high proportion of invalid programs. Second, the \textbf{performance landscape is highly non-linear and combinatorial} (Figure~\ref{fig:insight}b), where meaningful improvements often require coordinated structural changes rather than local edits. Such changes typically involve sequences of interdependent modifications, making kernel runtime optimization inherently a long-horizon search problem. In practice, expert programmers address this challenge by iteratively exploring alternative design choices, such as tiling strategies, memory layouts, and parallelization schemes, often requiring multiple rounds of trial-and-error refinement to achieve high-performance implementations~\citep{lim2017autotuning}.
In this work, we view kernel runtime optimization as a \textit{search problem under correctness constraints, with a highly non-linear and combinatorial performance landscape}. Rather than relying on external data or fine-tuning, we study how coding agents can progressively improve through accumulated execution feedback and structured exploration. These observations highlight two core challenges:
\textbf{(C1)} \textit{Feasible exploration}: how to ensure that exploration remains within the narrow feasible set defined by correctness constraints;
\textbf{(C2)} \textit{Optimization efficiency}: how to balance global exploration and local refinement in a rugged optimization landscape~\citep{sutton1998reinforcement}.

These challenges naturally suggest a decomposition:
(i) learning reusable constraints to stay within the feasible set, and
(ii) maintaining diverse candidates to explore the rugged landscape.
To this end, we propose \textbf{\METHOD{}}, an LLM-based kernel runtime optimization framework built on two complementary mechanisms: \textbf{\underline{Ada}ptation} and \textbf{\underline{Explor}ation}.
In the \textbf{adaptation} stage, the agent synthesizes training tasks and uses execution failures to construct a cross-task memory of validity rules that guide future generations toward the feasible set. Empirically, this substantially improves correctness on unseen problem instances and generalizes across different language models.
In the \textbf{exploration} stage, the agent performs a structured search on a tree of candidate kernels, maintaining multiple candidates, and exploring diverse solution trajectories as contextual signals. It alternates between small local edits and larger structural changes while reusing strong past candidates, enabling effective exploration beyond local optima. This yields improved inference-time scaling compared to common test-time optimization strategies, including iterative refinement, parallel sampling, and OpenEvolve~\citep{openevolve}, with gains increasing as the search budget grows (Figure~\ref{fig:ttt_summary}).
Together, these mechanisms help \METHOD{} reach valid kernels more reliably and search the optimization landscape more effectively for high-performing ones.
Our contributions are summarized as follows:
\begin{itemize}
    \item An analysis showing that kernel optimization is jointly bottlenecked by a sharp feasibility boundary (C1) and a highly non-smooth optimization landscape (C2).
    \item Two complementary mechanisms targeting these bottlenecks: a failure-driven memory that extracts reusable constraints to improve validity without model training, and a diversity-preserving structured search that balances local refinement and structural exploration for efficient test-time scaling.
    \item We show that the two mechanisms are complementary rather than redundant. On KernelBench, \METHOD{} reaches 3.11$\times$ speedup on Level-2 and 1.62$\times$ on Level-3 under a 100-step budget with GPT-5-mini as the base model.
\end{itemize}

\begin{figure*}[t]
    \centering
    \vspace{-10mm}
    \includegraphics[width=1\linewidth]{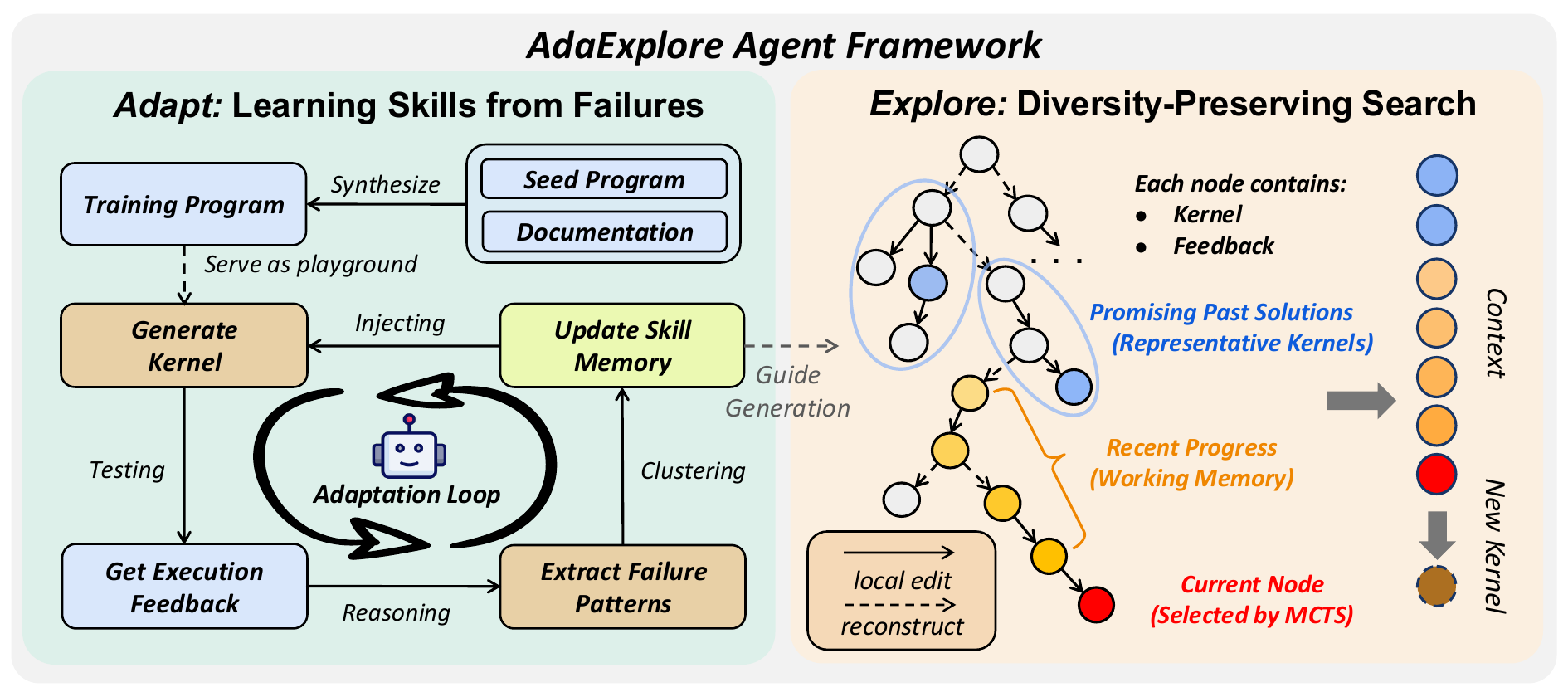}
    \vspace{-5mm}
    \caption{\textbf{Overview of \METHOD{} for Kernel Runtime Optimization.} The method has two stages: \textbf{Adapt}: it turns failures on synthesized tasks into a cross-task memory that helps generate correct kernels. \textbf{Explore}: it organizes candidate kernels as a tree and alternates between local refinement and regeneration to search for higher-performing solutions.}
    \vspace{-2mm}
    \label{fig:model}
\end{figure*}
\section{Related Work}
\textbf{LLMs for Code Generation.}
General-purpose code models \citep{openai_codex, li2022competition, roziere2023code, seed2025seed, hui2024qwen2} have achieved strong results on benchmarks such as HumanEval~\citep{chen2021evaluating} and SWE-bench~\citep{jimenez2023swe}, especially for well-represented languages with abundant training data. For multilingual code generation, benchmarks such as Multi-SWE-bench~\citep{zan2025multi} and HumanEval-XL~\citep{peng2024humaneval} suggest that current LLMs still struggle on low-resource or domain-specific languages.

\textbf{Self-Improving Code Agents.}
Recent self-improving code agents fall primarily into two categories: cross-task adaptation and within-task improvement. Cross-task methods accumulate reusable reflections, skills, or prompts from prior trials~\citep{shinn2023reflexion, zhao2023expel, Yuksekgonul2024TextGradA, Zhang2025AgenticCE, wang2023voyager, agrawal2026gepa}. For example, ExpeL~\citep{zhao2023expel} builds an offline experience pool by contrasting success/failure trajectories. Within-task methods iteratively refine or search candidate programs for single tasks using environment feedback, tracing back to genetic programming~\citep{koza1992gp} and including LLM-guided evolutionary search methods such as FunSearch~\citep{romera2024funsearch}, AlphaEvolve~\citep{novikov2025alphaevolve}, and CodeEvolve~\citep{assumpcao2025codeevolve}, as well as co-evolutionary and reward-driven methods such as CoCoEvo~\citep{li2025cocoevo}. Recent work also studies structured search as a form of inference-time scaling~\citep{snell2024scaling,light2024scattered,jiang2025aide}. Our method combines these two perspectives by distilling transferable failure patterns across tasks before search, then using diversity-preserving optimization within each task.

\textbf{LLM Agents for GPU Kernel Generation.}
We focus on Triton DSL, introduced by~\citet{tillet2019triton} as a Python-like language for developing GPU kernels. Benchmarks such as TritonBench~\citep{li2025tritonbench}, KernelBench~\citep{ouyang2025kernelbench}, and Flashinfer-Bench~\citep{xing2026flashinfer} show that current language models still struggle with GPU kernel tasks. Recent methods explore both agentic search and training. Astra~\citep{wei2025astramultiagentgpukernel} decomposes optimization into multiple agent roles, while AccelOpt~\citep{zhang2025acceloptselfimprovingllmagentic} combines beam search with an optimization memory for emerging AI accelerators. Concurrent work KernelSkill~\citep{sun2026kernelskill} builds reusable skills via expert analysis, whereas we automatically distill failure patterns from execution feedback. On the training side, Kevin~\citep{baronio2025kevinmultiturnrlgenerating} applies multi-turn GRPO, CUDA-L1~\citep{li2026cudal1improvingcudaoptimization} uses contrastive reinforcement learning with speedup-scored exemplars, and CUDA Agent~\citep{dai2026cudaagentlargescaleagentic} scales PPO with synthesized data. Our method is most related to this line of work, but it differs in explicitly combining reusable knowledge distillation with diversity-preserving optimization.

\section{Method}

\subsection{Task Setup}

We formulate kernel runtime optimization as a \textbf{program rewriting and optimization problem}.
\textit{The input} is a high-level implementation (e.g., Python) of an atomic function (e.g., matrix multiplication); \textit{the output} is a kernel written in one specific low-level language (e.g., CUDA/Triton) that \textit{(i) preserves functional correctness} and \textit{(ii) maximizes runtime performance on target hardware.} This makes the task inherently difficult: the agent must preserve the semantics of the high-level program while discovering low-level implementations that satisfy hardware constraints and achieve strong performance.

\subsection{Method Overview}

Our framework consists of two components: \textsc{Adapt} and \textsc{Explore}.
\textbf{Adapt} directly addresses feasible-set exploration by learning reusable constraints, thereby keeping the optimization process within the feasible set.
By running the agent on synthesized tasks and collecting execution failures, we build a compact cross-task skill memory of simple rules about what tends to invalidate kernels.
This cross-task skill memory reduces syntax and execution errors during inference, thus improving generation accuracy and implicitly accelerating search speed.
\textbf{Explore} helps search the optimization landscape more effectively for high-performing kernels by balancing candidate diversity and performance.
We keep candidate kernels in a tree rather than a single chain, so the search can preserve multiple promising directions at once.
Each expansion alternates between small local refinements and larger structural changes, using recent process together with previously discovered strong kernels to navigate the optimization landscape.
Detailed algorithms for both components are provided in Appendix~\ref{app:method_details}.


\subsection{Adapt: Learning Skills from Failures}

Empirically, kernel generation failures often \textit{stem from a small set of recurring grammar errors and structural constraints} (e.g., unsupported Triton operations, constexpr violations). Rather than relying on additional training, we adapt the model's knowledge about these constraints through self-exploration. As shown in Figure~\ref{fig:model} (left), we synthesize reference programs as training tasks and ask the agent to implement their corresponding kernel implementations. By summarizing the resulting execution feedback, we cluster recurring failure patterns and distill them into system instructions as a \textit{cross-task skill memory}.

\paragraph{Task Synthesis} Our pipeline starts from high-level reference implementations (e.g., PyTorch programs) composed of standard operators. We use a small set of task examples that differ from the test sets as seeds, and then reuse and recombine operators described in the language's operator documentation to continually synthesize diverse training tasks, rather than relying on a fixed, hand-curated set. For each synthesized task, the agent generates a corresponding low-level kernel and executes it against the reference implementation, so that failures reveal reusable validity constraints. This allows us to generate infinite and varied playgrounds for agents. The details of the implementation and statistics of the synthetic training set are provided in the Appendix~\ref{app:dataset_synthesis}.

\paragraph{Cross-Task Memory for Constraint-Aware Skills}
We use these synthesized tasks to explore constraints and extract cross-tasks from failed attempts (see Table~\ref{tab:memory} for examples). We maintain a lightweight cross-task skill memory that stores guidance such as avoiding incorrect function calls or common implementation pitfalls. Importantly, this memory is constructed in an evolving online manner: as we iterate over synthesized training tasks, newly extracted skills are continuously added to the memory, and the accumulated memory is exposed to subsequent tasks. This allows the method to transfer experience across tasks and avoid repeatedly falling into previously observed failure modes.
To construct the cross-task skill memory, we run coding agents on synthesized tasks and collect failed generations with their execution feedback. Each failure is summarized into a concise constraint rule (e.g., \textit{`you cannot generate a Triton pointer type inside a vectorized load'}), converting raw diagnostics into actionable guidance.

We then aggregate these rules by extracting recurring patterns using an LLM judge and recording their frequencies. Frequency serves as a proxy for generality: high-frequency rules capture common failure modes and provide broadly reusable guidance, while low-frequency rules often correspond to noise or task-specific edge cases. We therefore retain only rules whose frequency exceeds a threshold $O$ (set to 3 in practice), resulting in a compact and robust memory of reusable constraints.



\subsection{Explore: Diversity-Preserving Search}

Kernel optimization is inherently a long-horizon iterative process: the agent repeatedly proposes code edits or structural changes, executes the resulting kernels, and uses performance feedback to guide subsequent decisions.
However, incorporating the full trajectory into the prompt quickly exhausts the model's context budget, whereas aggressive truncation removes information about the search progress. In addition, overreliance on previous solutions can bias the model towards nearby variants and reduce the diversity of newly generated candidates~\citep{chu2024exploring}.

To address this, we introduce \textbf{Explore} (Figure~\ref{fig:model}, right), which builds on standard tree search to explore multiple candidate kernels beyond a single refinement chain.
We focus on two practical design choices:
(i) structuring the search tree and action space to support both local refinement and structural regeneration (Section~\ref{sec:tree_search}), and
(ii) constructing context by combining recent branch history with high-performing past candidates, together with an appropriate node selection strategy, to encourage diversity during search (Section~\ref{sec:context_mem}).

\subsubsection{Tree Search and Action Space}
\label{sec:tree_search}
We organize optimization as a search tree $\mathcal{T}$ rather than a single refinement chain. Each node $s \in \mathcal{T}$ is a kernel candidate, and expanding a node produces a new child candidate after execution feedback is observed. Tree search allows maintaining multiple feasible but structurally distinct candidates, which is critical in a complex search space. In contrast, a single chain can let early design decisions constrain all later refinements and limit global exploration.

\textbf{Action Space.}
At each step, from a selected node, the agent applies one of two update operators. \textbf{Small step} performs localized patch-based refinement, preserving the overall kernel structure while correcting errors or tuning local choices. \textbf{Large step} regenerates the kernel at a structural level, encouraging alternative strategies and broader exploration. A detailed description of the actions is in Appendix~\ref{app:actions}.

\subsubsection{Context and Node Selection}
\label{sec:context_mem}
\textbf{Context Management.} When expanding a node $s$, the model conditions on two sources of context: a \emph{working memory} taken from a limited recent window $C_{\text{recent}}$ along the path from the root to $s$, which stores recent edits and execution feedback, and a pool of \emph{representative kernels} extracted from earlier search stages. The working memory supports local correction, while the representative kernels preserve longer-horizon search signals without overloading the context. For a \textbf{large step}, we clear the working memory so that the agent can better break away from the current refinement chain. For a \textbf{small step}, by contrast, the model relies only on the local working memory, which keeps the update tightly grounded in the current branch and encourages incremental refinement. Together, this dual-memory design balances local refinement with broader exploration within the current search trajectory.

To avoid representing near-duplicate kernels, we partition the path into connected segments of consecutive small-step refinements and allow each segment to contribute at most one representative kernel. Let $\mathcal{K}_{\text{pool}}=\{k_1, \dots, k_{|\mathcal{K}_{\text{pool}}|}\}$ denote the resulting set of representative kernels for the current node $s$. We then uniformly sample at most $C_{\text{pool}}$ kernels.

\textbf{Node Selection.} We select the next expansion with a UCT-style rule~\citep{kocsis2006bandit} over existing children together with an explicit \emph{expand} option. For an existing child $a$ of node $s$, we use
\(
\mathrm{UCT}(s,a) = Q(s,a) + c_{\mathrm{explore}} \sqrt{\frac{\ln N(s)}{N(s,a)}},
\)
and for creating a new child, we use
\(
\mathrm{Expand}(s) = Q_{\mathrm{expand}}(s) + c_{\mathrm{expand}} \sqrt{\frac{\ln N(s)}{|\mathcal{C}(s)|^2}},
\)
where $Q(s,a)$ is the observed value of child $a$, $N(s,a)$ and $N(s)$ are visit counts, $|\mathcal{C}(s)|$ is the current number of children, and $Q_{\mathrm{expand}}(s) = \max_{a \in \mathcal{C}(s)} Q(s,a)$ is the best observed value among existing children. The coefficient $c_{\mathrm{explore}}$ controls the exploration-exploitation trade-off. This explicit \emph{expand} option reflects that the set of possible refinements or regenerations is not known in advance.

\section{Experiments}
\label{sec:experiments}
\subsection{Baselines}
We evaluate representative baselines that reflect common paradigms in kernel runtime optimization workflows.

\textbf{Single-Pass Baselines.}
We report single-pass results for GPT-5-mini, GPT-5~\citep{singh2025openai}, and Claude-4.6-Opus~\citep{anthropic2026claude46}, which reflect strong one-shot performance on kernel runtime optimization tasks. We also include AutoTriton~\citep{li2025autotriton}, a Triton-specialized model trained using execution-based rewards.

\textbf{Parallel-Sampling (PS).}
The LLM generates diverse candidate kernels in parallel and selects the best-performing one. We evaluate both the vanilla baseline and its variant augmented with cross-task skill memory \textit{(w. SM)}.

\textbf{Iterative-Refinement (IR).}
Starting from an initial kernel, the agent iteratively patches the previous candidate using compiler and runtime feedback, such as errors and running time. We evaluate both the vanilla baseline and its variant with cross-task skill memory \textit{(w. SM)}.

\textbf{DR. Kernel~\citep{liu2026dr}.}
An RL baseline for Triton kernel runtime optimization that combines multi-turn training with execution feedback and sequential test-time scaling. We report both single-pass results and best-performing results under a matched test-time budget
(4 samples $\times$ 14 steps), aligned with the scaling setup described in the paper.

\textbf{OpenEvolve~\citep{openevolve}.}
An open-source evolutionary coding agent that maintains a diverse population of candidate programs. This baseline represents a population-based search paradigm for code optimization.

\subsection{Testbeds and Metrics}
We use KernelBench~\citep{ouyang2025kernelbench} as our main testbed, which contains human-collected kernel runtime optimization tasks organized by difficulty: Level-1 covers single operators, Level-2 covers simple fused kernels, and Level-3 covers more complex model-level workloads.
Level-1 tasks in KernelBench are used for data synthesis, and we evaluate Level-2 and Level-3 tasks.
We additionally evaluate on FlashInfer-Bench~\citep{xing2026flashinfer}, whose kernel tasks are extracted from production LLM inference pipelines with real deployment shapes and expert-written FlashInfer CUDA kernels as strong baselines (Appendix~\ref{app:flashinfer_bench}).
In performance comparison, all kernels are executed and profiled on an NVIDIA A6000 GPU at a fixed frequency (1500 MHz).
The agent generates Triton kernels to accelerate the reference PyTorch programs provided in each task.
For all multi-pass baseline baselines except DR. Kernel, we use GPT-5-mini as the base model.
We use the following metrics:

\textbf{Accuracy:} The percentage of runs that produce at least one functionally correct kernel.

\textbf{Speedup:} The ratio between the runtime of the reference PyTorch eager implementation and that of the best among the generated kernels. We clip the speedup under $10$ to remove extreme outliers. We note that prior work often reports uncapped averages, which can be dominated by a small number of extreme cases, making comparisons less reliable. 

\textbf{Fast@$p$:} The percentage of runs that produce at least one correct kernel achieving a speedup greater than $p\times$ over the PyTorch eager baseline. We use $p=1.2$ to indicate a non-trivial improvement over the baseline and $p=2$ to indicate a significant improvement.

We measure kernel execution time using CUDA events. Each kernel undergoes 10 warm-up iterations followed by 100 timed trials. To reduce noise, we apply symmetric outlier trimming, discarding the fastest and slowest 5\% \ of the measurements, and computing statistics over the remaining 90 runs. Unless specified, all agents have a test-time budget of 50 steps. Detailed hyperparameters are listed in Appendix~\ref{app:openevolve_hyper}.

\begin{table}[t]
\centering
\vspace{-6mm}
\small
\setlength{\tabcolsep}{4pt}
\renewcommand{\arraystretch}{1.1}
\resizebox{\columnwidth}{!}{%
\begin{tabular}{l|cccc|cccc}
\toprule
\multirow{2}{*}{\textbf{Method}} 
& \multicolumn{4}{c|}{\textbf{KernelBench Level-2}} 
& \multicolumn{4}{c}{\textbf{KernelBench Level-3}} \\
\cmidrule(lr){2-5}\cmidrule(lr){6-9}
& \textbf{Acc.} & \textbf{Speedup} & \textbf{Fast@1.2} & \textbf{Fast@2}
& \textbf{Acc.} & \textbf{Speedup} & \textbf{Fast@1.2} & \textbf{Fast@2} \\
\midrule

\rowcolor{blue!10}\multicolumn{9}{l}{\textbf{\textit{Single-Pass Baselines}}} \\

\textbf{GPT-5-mini} 
& 22\% & 0.34$\times$ & 9\% & 3\% & 22\% & 0.21$\times$ & 2\% & 0\% \\
\textbf{GPT-5} 
& 51\% & 0.78$\times$ & 19\% & 4\% & 44\% & 0.55$\times$ & 12\% & 4\% \\

\textbf{Claude-4.6-Opus} 
& 60\% & 0.76$\times$ & 16\% & 5\% & 72\% & 0.84$\times$ & 8\% & 6\% \\

\textbf{AutoTriton} 
& 41\% & 0.44$\times$ & 7\% & 1\% & 32\% & 0.30$\times$ & 0\% & 0\% \\

\textbf{DR. Kernel} 
& 46\% & 0.86$\times$ & 14\% & 6\% & 10\% & 0.21$\times$ & 2\% & 0\% \\

\midrule

\rowcolor{blue!10}\multicolumn{9}{l}{\textbf{\textit{Multi-Pass Baselines}}} \\

\textbf{PS (50 Steps)} 
& 87\% & 1.69$\times$ & 49\% & 17\% & 92\% & 0.97$\times$ & 18\% & 4\% \\

\textbf{PS \textit{w. SM} (50 Steps)} 
& 100\% & 2.12$\times$ & 63\% & 23\%
& 98\% & 1.11$\times$ & \textbf{28\%} & 4\% \\

\textbf{IR (50 Steps)} 
& 100\% & 1.96$\times$ & 44\% & 20\%
& 100\% & 1.29$\times$ & \underline{24\%} & \underline{12\%} \\

\textbf{IR \textit{w. SM} (50 Steps)} 
& 100\% & \underline{2.59$\times$} & \underline{67\%} & \underline{33\%}
& 100\% & 1.16$\times$ & 10\% & 4\% \\

\textbf{OpenEvolve \textit{w. SM} (50 Steps)}
& 100\% & 1.91$\times$ & 35\% & 12\%
& 100\% & \underline{1.47$\times$} & \textbf{28\%} & 10\% \\

\textbf{DR. Kernel (4 $\times$ 14 steps)} 
& 100\% & 1.78$\times$ & 51\% & 13\%
& 84\% & 0.97$\times$ & 16\% & 8\% \\
\textbf{\METHOD{} (50 Steps) (Ours)} 
& 100\% & \textbf{2.63$\times$} & \textbf{68\%} & \textbf{34\%} & 100\% & \textbf{1.56$\times$} & \textbf{28\%} & \textbf{16\%} \\
\midrule
\rowcolor{gray!20}
\textbf{\METHOD{} (100 Steps) (Ours)} 
& 100\% & 3.11$\times$ & 77\% & 44\% & 100\% & 1.62$\times$ & 32\% & 18\% \\
\rowcolor{gray!20}
\textbf{\METHOD{} (200 Steps) (Ours)} 
& 100\% & 3.27$\times$ & 79\% & 48\% & 100\% & 1.72$\times$ & 36\% & 22\% \\
\bottomrule
\end{tabular}
}
\caption{\textbf{Performance Comparison.} We report Acc., Speedup, Fast@1.2, and Fast@2. \textit{w. SM} denotes augmenting with our cross-task skill memory.
Among entries other than Acc., the best result in multi-pass baselines is highlighted in \textbf{bold}, and the second-best result is \underline{underlined}. The shaded row reports performance with increased compute (100 steps), illustrating scaling behavior. GPT-5-mini is the default base model for multi-pass baselines.}
\label{tab:main_results_50}
\end{table}
\begin{table}[t]
\centering
\vspace{-1mm}
\small
\begin{tabular}{l|cc|cc|c|c}
\toprule
\textbf{Setting}
& \multicolumn{2}{c|}{\textbf{GPT-5-mini}}
& \multicolumn{2}{c|}{\textbf{Qwen3-Coder-Next}}
& \multicolumn{1}{c|}{\textbf{GPT-5}}
& \multicolumn{1}{c}{\textbf{Claude-4.6-Opus}} \\
\cmidrule(lr){2-3} \cmidrule(lr){4-5} \cmidrule(lr){6-7}
& Pass@1 & Pass@25 & Pass@1 & Pass@25 & Pass@1 & Pass@1 \\
\midrule
\textit{Direct}
& 22\% & 76\% & 7\%  & 58\% & 51\% & 60\% \\
\textit{+ Static Docs}
& 34\% &  94\% &  11\% & 69\% & 60\% & 67\% \\
\textit{+ Skill Memory (Ours)}
& \textbf{54\%} & \textbf{100\%} & \textbf{17\%} & \textbf{88\%} & \textbf{67\%} & \textbf{72\%} \\
\bottomrule
\end{tabular}
\caption{\textbf{KernelBench L2 Accuracy with Skill Memory.} Self-generated cross-task skill memory improves correctness, and the mechanism can be applied to various models.}
\label{tab:knowledge_base_performance}
\end{table}

\subsection{Results}

\textbf{Comparison of \METHOD{} with the Baselines.}
Table~\ref{tab:main_results_50} compares single-pass and multi-pass baselines on KernelBench Level-2 and Level-3. Single-pass frontier models struggle on Level-2: GPT-5-mini reaches only 22\% correctness, GPT-5 reaches 51\%, and Claude-4.6-Opus reaches 60\%. This gap highlights the difficulty of generating correct kernels.

Under multi-pass test-time optimization, most methods achieve high correctness at Level-2, but their optimization quality varies substantially. \METHOD{} reaches \textbf{100\% correctness, 2.63$\times$ speedup, 68\% Fast@1.2 and 34\% Fast@2} at step 50, on par with the strongest non-\METHOD{} baseline, IR \textit{w.\ SM}, which augments iterative refinement with our cross-task skill memory. They separate once the budget grows, with \METHOD{} reaching 3.11$\times$ against 2.88$\times$ at step 100 (see Figure~\ref{fig:ttt_summary}, left). The result suggests that our skill memory is beneficial in all search strategies and that combining with \textbf{Explore} yields even stronger optimization.
The same trend largely carries over to the harder Level-3 setting. \METHOD{} again achieves the best performance at step 50, with \textbf{1.56$\times$ speedup, 28\% Fast@1.2 and 16\% Fast@2}. 
Additionally, we find that \METHOD{} transfers well across GPU generations with the same cross-task skill memory, as shown in Table~\ref{tab:gpu_result}.

\textit{IR w. SM} shows a clear Level-2/Level-3 gap. 
L2 tasks often center on optimizing a single kernel, where local edits are effective; in contrast, L3 tasks involve model-level structures such as ResNet and LSTM, where performance depends on higher-level design choices. 
Thus, refinement can remain trapped in local regions, with skill memory further encouraging conservative updates, motivating our structured search with large-step actions.


We further evaluate \METHOD{} on FlashInfer-Bench~\citep{xing2026flashinfer} by case study (Details in Appendix~\ref{app:flashinfer_bench}). On RMSNorm, the best generated kernel achieves \textbf{7.22$\times$} over PyTorch and \textbf{1.75$\times$} over the expert FlashInfer CUDA kernel; on GQA paged decode, it reaches \textbf{18.17$\times$} over PyTorch.
These results show that LLM agents have the potential to beat PyTorch and expert-written kernels, but compute-intensive, hardware-specialized kernels (e.g., those using Blackwell-specific instructions) remain challenging to beat.
\begin{table}[t]
\vspace{-6mm}
\centering
\small
\begin{minipage}[t]{0.5\linewidth}
\vspace{10pt}
\centering
\begin{tabular}{l r}
\toprule
\textbf{Category} & \textbf{Count} \\
\midrule
Training set size & 200 \\
Experience collected & 1178 \\
Deduplicated experience & 174 \\
Selected experience ($O \geq 3$) & 78 \\
\bottomrule
\end{tabular}
\end{minipage}
\hfill
\begin{minipage}[t]{0.40\linewidth}
\vspace{-6pt}
\centering
\includegraphics[trim=0 6 0 12,clip,width=\linewidth]{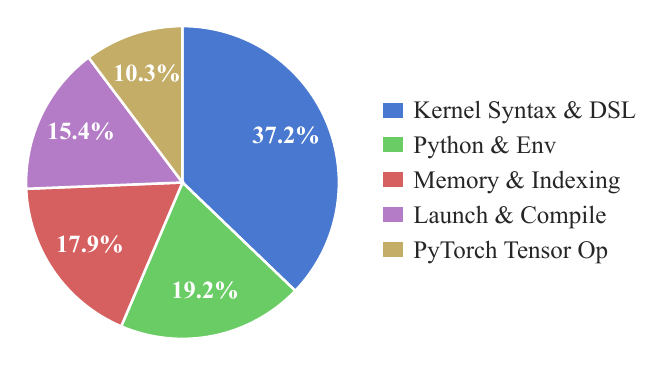}
\end{minipage}
\vspace{-2mm}
\caption{\textbf{Cross-Task Skill Memory Statistics.} From 200 synthesized tasks, 1,178 raw experiences are reduced to 174 unique items and 78 retained high-frequency hints ($O \geq 3$). The pie chart summarizes their category distribution.}
\label{tab:cross_problem_memory}
\vspace{-3mm}
\end{table}

\begin{wraptable}{r}{0.50\linewidth}
\vspace{-2mm}
\centering
\footnotesize
\setlength{\tabcolsep}{4pt}
\renewcommand{\arraystretch}{1.05}
\begin{tabular}{lccc}
\toprule
\textbf{GPU Version}  & \textbf{L40S} & \textbf{A100}  & \textbf{GB200} \\
\textbf{Architecture}
& Ada & Ampere & Blackwell \\
\midrule
\textbf{Speedup}
&  2.13$\times$ & 3.07$\times$ & 2.98$\times$ \\
\textbf{Correctness}
&  99\% & 100\% & 100\% \\
\bottomrule
\end{tabular}
\caption{\textbf{Cross-GPU Generalization of \METHOD{}.} We report speedup on KernelBench Level-2 tasks within 50 steps. The cross-task skill memory is collected on the A6000 but may transfer to other GPU versions.}
\label{tab:gpu_result}
\vspace{-2mm}
\end{wraptable}

\textbf{The Effectiveness of Skill Memory.}
Table~\ref{tab:knowledge_base_performance} isolates the effect of cross-task skill memory on KernelBench Level-2 in multiple model families, including GPT-5, Qwen3-Coder-Next~\citep{cao2026qwen3}, and Claude-4.6-Opus, and compares skill memory with static documentation (See Appendix~\ref{app:static_doc}). After each base model adapts its own skill memory, correctness improves consistently and more strongly than using static documentation. For example, for GPT-5-mini, Pass@1 increases from 22\% to 54\%, and Pass@25 increases from 76\% to 100\%.
These gains suggest that the adaptation procedure is broadly effective across models: each model can distill useful constraints from its own failure patterns and then improve generation correctness. 
To test whether cross-task skill memory can generalize across different benchmarks, we test the same skill memory on another benchmark, TritonBench, and also find a $28\%$ accuracy improvement (see Appendix~\ref{app:tritonbench} for details). We further report the full \METHOD{} pipeline on the small open-weight Qwen3-Coder-Next (Appendix~\ref{app:qwen_pipeline}) and a cross-model memory-transfer study (Appendix~\ref{app:cross_model}).

\textbf{Skill Memory Statistics.}
Table~\ref{tab:cross_problem_memory} shows that the memory extracted from 200 synthesized tasks is highly compact: 1,178 raw experiences are reduced to 174 deduplicated rules and finally 78 high-frequency transferable hints. This redundancy suggests that kernel-generation failures concentrate around a small set of recurring constraints, which helps explain why the resulting memory generalizes across tasks and models. We provide further details and examples of retained memory in Appendix~\ref{app:memory_analysis}.

\textbf{Test-Time Scaling of \METHOD{}.}
Figure~\ref{fig:ttt_summary} (left) illustrates the scaling behavior of the test time of \METHOD{} compared to the baselines as the inference budget increases by up to 200 steps. Throughout the trajectory, \METHOD{} consistently achieves a \textbf{higher best speedup}. Notably, its performance continues to improve without clear signs of saturation, and the gap relative to baselines widens further as more compute is allocated ($>50$ steps). This suggests that \METHOD{} retains substantial efficiency for further gains under larger inference budgets. Additionally, iterative refinement with our skill memory achieves comparable performance in the moderate-compute regime ($\sim 25$ steps). One interesting observation along the way is that the kernels produced by iterative refinement grow steadily in size as the budget increases while ours do not, suggesting that refinement increasingly patches an existing implementation rather than restructuring it (Appendix~\ref{app:kernel_growth}).



\begin{figure}[t]
    \centering
    \vspace{-6mm}
    \begin{minipage}[t]{0.45\linewidth}
        \vspace{0pt}
        \centering
        \includegraphics[
            width=\linewidth,
            trim=7.3 3 8.8 7,
            clip
        ]{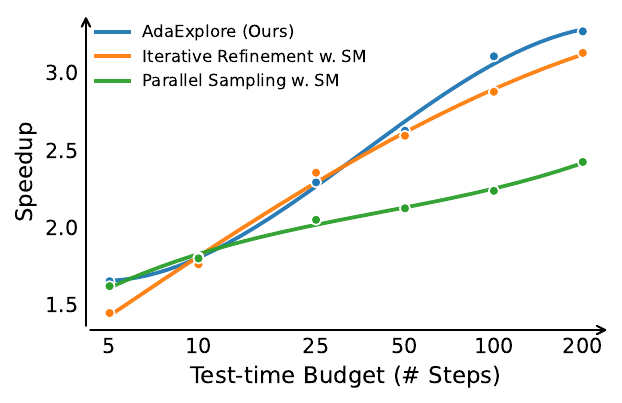}
    \end{minipage}
    \hfill
    \begin{minipage}[t]{0.5\linewidth}
        \vspace{0pt}
        \centering
        \footnotesize
        \setlength{\tabcolsep}{3pt}
        \renewcommand{\arraystretch}{0.95}
        \begin{tabular}{c c c l}
        \toprule
        Step & Type & Speedup & Code Changes \\
        \midrule
        1  & initial & 0.70 & +122  \\
        2  & small   & 0.85 & +25/--23 (71.97\%) \\

        \rowcolor{blue!10}
        5  & large   & 1.91 & +70/--61 (\textbf{19.81\%}) \\

        6  & small   & 2.27 & +64/--37 (61.38\%) \\
        8  & small   & 2.33 & +36/--23 (88.45\%) \\
        11 & small   & 2.42 & +27/--15 (82.40\%) \\

        \rowcolor{blue!10}
        20 & large   & 3.61 & +112/--119 (\textbf{20.17\%}) \\

        21 & small   & 3.72 & +29/--11 (85.15\%) \\
        22 & small   & 3.73 & +48/--25 (75.44\%) \\
        \bottomrule
        \end{tabular}%
    \end{minipage}

    \caption{\textbf{Test-time Scaling and Case Study on Actions.}
    \textbf{Left: }Average best-so-far speedup as the test-time budget increases, showing that \METHOD{} continues to efficiently improve with more search steps. \textbf{Right: } Case study illustrating the roles of large and small steps.}
    \label{fig:ttt_summary}
\end{figure}

\textbf{Case Study on Large and Small Steps.}
Figure~\ref{fig:ttt_summary} (right) illustrates a chain-only trajectory that isolates the roles of the two update types.
Large steps introduce low-similarity structural changes (19\%--20\%) and produce major speedup jumps, while small steps preserve higher similarity (60\%--88\%) and yield incremental gains.
Together, they allow \METHOD{} to combine structural exploration with stable local refinement.






\section{Ablation Study}

We ablate the main components of \METHOD{}: the tree-structured search design, the dual-action update space, the cross-task skill memory, and the representative kernel pool. Table~\ref{tab:ablation_main} summarizes the ablation settings and results.

\begin{wraptable}{r}{0.51\textwidth}
\centering
\footnotesize
\vspace{-3mm}
\setlength{\tabcolsep}{4pt}
\renewcommand{\arraystretch}{1.1}
\begin{tabular}{@{}lccc@{}}
\toprule
\textbf{Variant} & \textbf{Acc.} & \textbf{Speedup} & \textbf{Fast@1.2} \\
\midrule
AE w/o Small Step & 99\% & 2.35$\times$ & 60\% \\
AE w/o Large Step & 100\% & 2.60$\times$ & \textbf{69\%} \\
AE w/o Skill Memory & 99\% & 2.23$\times$ & 55\% \\
AE w/o Rep. Kernel & 100\% & 2.30$\times$ & 63\%\\
AE w/o MCTS & 100\% & 2.45$\times$ & 64\% \\
\midrule
\textbf{\METHOD{} (Full)} & 100\% & \textbf{2.63$\times$} & \textbf{69\%} \\
\bottomrule
\end{tabular}
\caption{\textbf{Ablation Study on \METHOD{} Components.} We separately remove action types, cross-task skill memory, representative kernels (Rep. Kernel), and tree search (MCTS) to isolate each component.}
\label{tab:ablation_main}
\vspace{-3mm}
\end{wraptable}

\textbf{Search Structure: Tree Search vs. Chain Search.}
We construct \METHOD{} without MCTS, which uses the same context management and action space as \METHOD{} but restricts the search to a chain without branching. The \METHOD{} without MCTS achieves lower speedup (2.45$\times$ vs.\ 2.63$\times$), suggesting that keeping multiple branches alive yields a substantial benefit.

\textbf{Action Space: Small Step vs. Large Step.} 
\textbf{w/o Large Step}, using only localized patch refinement, and \textbf{w/o Small Step}, using only structural regeneration. w/o Large Step remains close to the full method (2.60 vs.\ 2.63$\times$), likely because branching already provides diversity. w/o Small Step drops to 99\% correctness and 2.35$\times$, showing that structural changes need local refinement to reliably achieve correctness and improve performance.

\textbf{Cross-Task Skill Memory.}
We remove the skill memory while keeping other components fixed (\textbf{w/o Skill Memory}), resulting in a large performance drop from 2.63$\times$ to 2.23$\times$.

\textbf{Representative Kernel Pool.} The representative kernel pool serves as long-term progress storage during the search stage, helping the system make continuous progress. Removing it reduces the resulting speedup from 2.63$\times$ to 2.30$\times$.


\section{Conclusion}
In this work, we study kernel runtime optimization as a setting where coding agents must both generate \textit{correct kernels} and \textit{search effectively for high-performing ones}. 
We propose \METHOD{}, which combines an adaptation stage that builds cross-task skill memory from compiler and runtime failures and an exploration stage that uses tree-structured search to explore beyond local edits.
Experiments show that skill memory improves correctness, while search finds higher-performing kernels once correctness is reached, providing a practical way to improve kernel runtime optimization without additional fine-tuning.


\ifcolmsubmission\else
\section*{Acknowledgements}
This work was supported in part by SoftBank Group Corp. and Arm. This program was made possible (in part) due to the generosity of SoftBank Group Corp. We thank our collaborators at Arm for their support and collaboration.
\fi

\bibliography{colm2026_conference}

@misc{openai_codex,
  author       = {OpenAI},
  title        = {Introducing Codex},
  howpublished = {\url{https://openai.com/index/introducing-codex/}},
  year         = {2025}
}

@article{li2022competition,
  title={Competition-level code generation with alphacode},
  author={Li, Yujia and Choi, David and Chung, Junyoung and Kushman, Nate and Schrittwieser, Julian and Leblond, R{\'e}mi and Eccles, Tom and Keeling, James and Gimeno, Felix and Dal Lago, Agustin and others},
  journal={Science},
  volume={378},
  number={6624},
  pages={1092--1097},
  year={2022},
  publisher={American Association for the Advancement of Science}
}

@article{roziere2023code,
  title={Code llama: Open foundation models for code},
  author={Roziere, Baptiste and Gehring, Jonas and Gloeckle, Fabian and Sootla, Sten and Gat, Itai and Tan, Xiaoqing Ellen and Adi, Yossi and Liu, Jingyu and Sauvestre, Romain and Remez, Tal and others},
  journal={arXiv preprint arXiv:2308.12950},
  year={2023}
}

@article{seed2025seed,
  title={Seed-coder: Let the code model curate data for itself},
  author={Seed, ByteDance and Zhang, Yuyu and Su, Jing and Sun, Yifan and Xi, Chenguang and Xiao, Xia and Zheng, Shen and Zhang, Anxiang and Liu, Kaibo and Zan, Daoguang and others},
  journal={arXiv preprint arXiv:2506.03524},
  year={2025}
}

@article{hui2024qwen2,
  title={Qwen2. 5-coder technical report},
  author={Hui, Binyuan and Yang, Jian and Cui, Zeyu and Yang, Jiaxi and Liu, Dayiheng and Zhang, Lei and Liu, Tianyu and Zhang, Jiajun and Yu, Bowen and Lu, Keming and others},
  journal={arXiv preprint arXiv:2409.12186},
  year={2024}
}

@article{chen2021evaluating,
  title={Evaluating large language models trained on code},
  author={Chen, Mark and Tworek, Jerry and Jun, Heewoo and Yuan, Qiming and Pinto, Henrique Ponde De Oliveira and Kaplan, Jared and Edwards, Harri and Burda, Yuri and Joseph, Nicholas and Brockman, Greg and others},
  journal={arXiv preprint arXiv:2107.03374},
  year={2021}
}

@article{jimenez2023swe,
  title={Swe-bench: Can language models resolve real-world github issues?},
  author={Jimenez, Carlos E and Yang, John and Wettig, Alexander and Yao, Shunyu and Pei, Kexin and Press, Ofir and Narasimhan, Karthik},
  journal={arXiv preprint arXiv:2310.06770},
  year={2023}
}

@article{zan2025multi,
  title={Multi-swe-bench: A multilingual benchmark for issue resolving},
  author={Zan, Daoguang and Huang, Zhirong and Liu, Wei and Chen, Hanwu and Zhang, Linhao and Xin, Shulin and Chen, Lu and Liu, Qi and Zhong, Xiaojian and Li, Aoyan and others},
  journal={arXiv preprint arXiv:2504.02605},
  year={2025}
}

@article{peng2024humaneval,
  title={Humaneval-xl: A multilingual code generation benchmark for cross-lingual natural language generalization},
  author={Peng, Qiwei and Chai, Yekun and Li, Xuhong},
  journal={arXiv preprint arXiv:2402.16694},
  year={2024}
}

@inproceedings{tillet2019triton,
  title={Triton: an intermediate language and compiler for tiled neural network computations},
  author={Tillet, Philippe and Kung, Hsiang-Tsung and Cox, David},
  booktitle={Proceedings of the 3rd ACM SIGPLAN International Workshop on Machine Learning and Programming Languages},
  pages={10--19},
  year={2019}
}

@article{li2025tritonbench,
  title={TritonBench: Benchmarking Large Language Model Capabilities for Generating Triton Operators},
  author={Li, Jianling and Li, Shangzhan and Gao, Zhenye and Shi, Qi and Li, Yuxuan and Wang, Zefan and Huang, Jiacheng and Wang, Haojie and Wang, Jianrong and Han, Xu and others},
  journal={arXiv preprint arXiv:2502.14752},
  year={2025}
}

@article{ouyang2025kernelbench,
  title={Kernelbench: Can llms write efficient gpu kernels?},
  author={Ouyang, Anne and Guo, Simon and Arora, Simran and Zhang, Alex L and Hu, William and R{\'e}, Christopher and Mirhoseini, Azalia},
  journal={arXiv preprint arXiv:2502.10517},
  year={2025}
}

@book{sutton1998reinforcement,
  title={Reinforcement learning: An introduction},
  author={Sutton, Richard S and Barto, Andrew G and others},
  volume={1},
  number={1},
  year={1998},
  publisher={MIT press Cambridge}
}

@article{guo2025deepseekr1,
  title={DeepSeek-R1: Incentivizing Reasoning Capability in LLMs via Reinforcement Learning},
  author={Guo, Daya and Yang, Dejian and Zhang, Haowei and Song, Junxiao and Zhang, Ruoyu and Xu, Runxin and Zhu, Qihao and Ma, Shirong and Wang, Peiyi and Bi, Xiao and others},
  journal={arXiv preprint arXiv:2501.12948},
  year={2025}
}

@misc{xu2025wizardlmempoweringlargepretrained,
      title={WizardLM: Empowering large pre-trained language models to follow complex instructions},
      author={Can Xu and Qingfeng Sun and Kai Zheng and Xiubo Geng and Pu Zhao and Jiazhan Feng and Chongyang Tao and Qingwei Lin and Daxin Jiang},
      year={2025},
      eprint={2304.12244},
      archivePrefix={arXiv},
      primaryClass={cs.CL},
      url={https://arxiv.org/abs/2304.12244},
}

@article{Yuksekgonul2024TextGradA,
  title={TextGrad: Automatic "Differentiation" via Text},
  author={Mert Yuksekgonul and Federico Bianchi and Joseph Boen and Sheng Liu and Zhi Huang and Carlos Guestrin and James Zou},
  journal={ArXiv},
  year={2024},
  volume={abs/2406.07496},
}

@article{Zhang2025AgenticCE,
  title={Agentic Context Engineering: Evolving Contexts for Self-Improving Language Models},
  author={Qizheng Zhang and Changran Hu and Shubhangi Upasani and Boyuan Ma and Fenglu Hong and Vamsidhar Reddy Kamanuru and Jay Rainton and Chen Wu and Mengmeng Ji and Hanchen Li and Urmish Thakker and James Zou and Kunle Olukotun},
  journal={ArXiv},
  year={2025},
  volume={abs/2510.04618},
}

@article{xing2026flashinfer,
  title={FlashInfer-Bench: Building the Virtuous Cycle for AI-driven LLM Systems},
  author={Xing, Shanli and Zhai, Yiyan and Jiang, Alexander and Dong, Yixin and Wu, Yong and Ye, Zihao and Ruan, Charlie and Huang, Yingyi and Zhang, Yineng and Yin, Liangsheng and others},
  journal={arXiv preprint arXiv:2601.00227},
  year={2026}
}

@article{nijkamp2022codegen,
  title={Codegen: An open large language model for code with multi-turn program synthesis},
  author={Nijkamp, Erik and Pang, Bo and Hayashi, Hiroaki and Tu, Lifu and Wang, Huan and Zhou, Yingbo and Savarese, Silvio and Xiong, Caiming},
  journal={arXiv preprint arXiv:2203.13474},
  year={2022}
}

@misc{shinn2023reflexion,
  title={Reflexion: Language Agents with Verbal Reinforcement Learning},
  author={Shinn, Noah and Cassano, Federico and Berman, Edward and Gopinath, Ashwin and Narasimhan, Karthik and Yao, Shunyu},
  year={2023},
  eprint={2303.11366},
  archivePrefix={arXiv},
  primaryClass={cs.AI},
  url={https://arxiv.org/abs/2303.11366}
}

@misc{zhao2023expel,
  title={ExpeL: LLM Agents Are Experiential Learners},
  author={Zhao, Andrew and Huang, Daniel and Xu, Quentin and Lin, Matthieu and Liu, Yong-Jin and Huang, Gao},
  year={2023},
  eprint={2308.10144},
  archivePrefix={arXiv},
  primaryClass={cs.AI},
  url={https://arxiv.org/abs/2308.10144}
}

@inproceedings{qian2024chatdev,
  title={Chatdev: Communicative agents for software development},
  author={Qian, Chen and Liu, Wei and Liu, Hongzhang and Chen, Nuo and Dang, Yufan and Li, Jiahao and Yang, Cheng and Chen, Weize and Su, Yusheng and Cong, Xin and others},
  booktitle={Proceedings of the 62nd annual meeting of the association for computational linguistics (volume 1: Long papers)},
  pages={15174--15186},
  year={2024}
}

@inproceedings{zhang2024codeagent,
  title={Codeagent: Enhancing code generation with tool-integrated agent systems for real-world repo-level coding challenges},
  author={Zhang, Kechi and Li, Jia and Li, Ge and Shi, Xianjie and Jin, Zhi},
  booktitle={Proceedings of the 62nd Annual Meeting of the Association for Computational Linguistics (Volume 1: Long Papers)},
  pages={13643--13658},
  year={2024}
}

@article{wang2025agents,
  title={Agents in software engineering: Survey, landscape, and vision},
  author={Wang, Yanlin and Zhong, Wanjun and Huang, Yanxian and Shi, Ensheng and Yang, Min and Chen, Jiachi and Li, Hui and Ma, Yuchi and Wang, Qianxiang and Zheng, Zibin},
  journal={Automated Software Engineering},
  volume={32},
  number={2},
  pages={70},
  year={2025},
  publisher={Springer}
}

@book{koza1992gp,
  author    = {John R. Koza},
  title     = {Genetic Programming: On the Programming of Computers by Means of Natural Selection},
  publisher = {MIT Press},
  address   = {Cambridge, MA, USA},
  year      = {1992}
}

@article{romera2024funsearch,
  author  = {Bernardino Romera-Paredes and Mohammadamin Barekatain and Alexander Novikov and Matej Balog and M. Pawan Kumar and Emilien Dupont and Francisco J. R. Ruiz and others},
  title   = {Mathematical discoveries from program search with large language models},
  journal = {Nature},
  volume  = {625},
  pages   = {468--475},
  year    = {2024}
}

@misc{novikov2025alphaevolve,
  author       = {Alexander Novikov and Ng{\^{a}}n V{\~u} and Marvin Eisenberger and Emilien Dupont and Po-Sen Huang and Adam Zsolt Wagner and Sergey Shirobokov and Borislav Kozlovskii and Francisco J. R. Ruiz and Abbas Mehrabian and M. Pawan Kumar and Abigail See and Swarat Chaudhuri and George Holland and Alex Davies and Sebastian Nowozin and Pushmeet Kohli and Matej Balog},
  title        = {AlphaEvolve: A coding agent for scientific and algorithmic discovery},
  howpublished = {arXiv preprint arXiv:2506.13131},
  year         = {2025}
}

@misc{assumpcao2025codeevolve,
  author       = {{Assump\c{c}\~{a}o}, Henrique and Ferreira, Diego and Campos, Leandro and Murai, Fabricio},
  title        = {CodeEvolve: an open source evolutionary coding agent for algorithm discovery and optimization},
  howpublished = {arXiv preprint arXiv:2510.14150},
  year         = {2025}
}

@misc{li2025cocoevo,
  author       = {Kefan Li and Hongyue Yu and Tingyu Guo and Shijie Cao and Yuan Yuan},
  title        = {CoCoEvo: Co-Evolution of Programs and Test Cases to Enhance Code Generation},
  howpublished = {arXiv preprint arXiv:2502.10802},
  year         = {2025}
}

@software{openevolve,
  title = {OpenEvolve: an open-source evolutionary coding agent},
  author = {Asankhaya Sharma},
  year = {2025},
  publisher = {GitHub},
  url = {https://github.com/algorithmicsuperintelligence/openevolve}
}

@article{liu2026dr,
  title={Dr. Kernel: Reinforcement Learning Done Right for Triton Kernel Generations},
  author={Liu, Wei and Xu, Jiawei and Li, Yingru and Zheng, Longtao and Li, Tianjian and Liu, Qian and He, Junxian},
  journal={arXiv preprint arXiv:2602.05885},
  year={2026}
}

@article{li2025autotriton,
  title={Autotriton: Automatic triton programming with reinforcement learning in llms},
  author={Li, Shangzhan and Wang, Zefan and He, Ye and Li, Yuxuan and Shi, Qi and Li, Jianling and Hu, Yonggang and Che, Wanxiang and Han, Xu and Liu, Zhiyuan and others},
  journal={arXiv preprint arXiv:2507.05687},
  year={2025}
}

@article{chu2024exploring,
  title={Exploring and controlling diversity in llm-agent conversation},
  author={Chu, KuanChao and Chen, Yi-Pei and Nakayama, Hideki},
  journal={arXiv preprint arXiv:2412.21102},
  year={2024}
}

@inproceedings{lim2017autotuning,
  title={Autotuning GPU kernels via static and predictive analysis},
  author={Lim, Robert and Norris, Boyana and Malony, Allen},
  booktitle={2017 46th international conference on parallel processing (icpp)},
  pages={523--532},
  year={2017},
  organization={IEEE}
}

@misc{dai2026cudaagentlargescaleagentic,
      title={CUDA Agent: Large-Scale Agentic RL for High-Performance CUDA Kernel Generation},
      author={Weinan Dai and Hanlin Wu and Qiying Yu and Huan-ang Gao and Jiahao Li and Chengquan Jiang and Weiqiang Lou and Yufan Song and Hongli Yu and Jiaze Chen and Wei-Ying Ma and Ya-Qin Zhang and Jingjing Liu and Mingxuan Wang and Xin Liu and Hao Zhou},
      year={2026},
      eprint={2602.24286},
      archivePrefix={arXiv},
      primaryClass={cs.LG},
      url={https://arxiv.org/abs/2602.24286},
}

@misc{baronio2025kevinmultiturnrlgenerating,
      title={Kevin: Multi-Turn RL for Generating CUDA Kernels},
      author={Carlo Baronio and Pietro Marsella and Ben Pan and Simon Guo and Silas Alberti},
      year={2025},
      eprint={2507.11948},
      archivePrefix={arXiv},
      primaryClass={cs.LG},
      url={https://arxiv.org/abs/2507.11948},
}

@misc{li2026cudal1improvingcudaoptimization,
      title={CUDA-L1: Improving CUDA Optimization via Contrastive Reinforcement Learning},
      author={Xiaoya Li and Xiaofei Sun and Albert Wang and Jiwei Li and Chris Shum},
      year={2026},
      eprint={2507.14111},
      archivePrefix={arXiv},
      primaryClass={cs.AI},
      url={https://arxiv.org/abs/2507.14111},
}

@misc{wei2025astramultiagentgpukernel,
      title={Astra: A Multi-Agent System for GPU Kernel Performance Optimization},
      author={Anjiang Wei and Tianran Sun and Yogesh Seenichamy and Hang Song and Anne Ouyang and Azalia Mirhoseini and Ke Wang and Alex Aiken},
      year={2025},
      eprint={2509.07506},
      archivePrefix={arXiv},
      primaryClass={cs.DC},
      url={https://arxiv.org/abs/2509.07506},
}

@misc{zhang2025acceloptselfimprovingllmagentic,
      title={AccelOpt: A Self-Improving LLM Agentic System for AI Accelerator Kernel Optimization},
      author={Genghan Zhang and Shaowei Zhu and Anjiang Wei and Zhenyu Song and Allen Nie and Zhen Jia and Nandita Vijaykumar and Yida Wang and Kunle Olukotun},
      year={2025},
      eprint={2511.15915},
      archivePrefix={arXiv},
      primaryClass={cs.LG},
      url={https://arxiv.org/abs/2511.15915},
}

@article{sun2026kernelskill,
  title={KernelSkill: A Multi-Agent Framework for GPU Kernel Optimization},
  author={Sun, Qitong and Han, Jun and Li, Tianlin and Tang, Zhe and Chen, Sheng and Yang, Fei and Liu, Aishan and Liu, Xianglong and Liu, Yang},
  journal={arXiv preprint arXiv:2603.10085},
  year={2026}
}

@article{snell2024scaling,
  title={Scaling llm test-time compute optimally can be more effective than scaling model parameters},
  author={Snell, Charlie and Lee, Jaehoon and Xu, Kelvin and Kumar, Aviral},
  journal={arXiv preprint arXiv:2408.03314},
  year={2024}
}

@article{light2024scattered,
  title={Scattered forest search: Smarter code space exploration with llms},
  author={Light, Jonathan and Wu, Yue and Sun, Yiyou and Yu, Wenchao and Zhao, Xujiang and Hu, Ziniu and Chen, Haifeng and Cheng, Wei and others},
  journal={arXiv preprint arXiv:2411.05010},
  year={2024}
}

@article{wang2023voyager,
  title={Voyager: An open-ended embodied agent with large language models},
  author={Wang, Guanzhi and Xie, Yuqi and Jiang, Yunfan and Mandlekar, Ajay and Xiao, Chaowei and Zhu, Yuke and Fan, Linxi and Anandkumar, Anima},
  journal={arXiv preprint arXiv:2305.16291},
  year={2023}
}

@article{cao2026qwen3,
  title={Qwen3-coder-next technical report},
  author={Cao, Ruisheng and Chen, Mouxiang and Chen, Jiawei and Cui, Zeyu and Feng, Yunlong and Hui, Binyuan and Jing, Yuheng and Li, Kaixin and Li, Mingze and Lin, Junyang and others},
  journal={arXiv preprint arXiv:2603.00729},
  year={2026}
}

@techreport{anthropic2026claude46,
  author = {Anthropic},
  title  = {Claude Opus 4.6 System Card},
  year   = {2026},
  institution = {Anthropic},
  url    = {https://www-cdn.anthropic.com/6a5fa276ac68b9aeb0c8b6af5fa36326e0e166dd.pdf}
}

@article{singh2025openai,
  title={{OpenAI GPT-5} system card},
  author={Singh, Aaditya and Fry, Adam and Perelman, Adam and Tart, Adam and Ganesh, Adi and El-Kishky, Ahmed and McLaughlin, Aidan and Low, Aiden and Ostrow, AJ and Ananthram, Akhila and others},
  journal={arXiv preprint arXiv:2601.03267},
  year={2026}
}

@inproceedings{kocsis2006bandit,
  title={Bandit based monte-carlo planning},
  author={Kocsis, Levente and Szepesv{\'a}ri, Csaba},
  booktitle={European conference on machine learning},
  pages={282--293},
  year={2006},
  organization={Springer}
}

@article{jiang2025aide,
  title={AIDE: AI-Driven Exploration in the Space of Code},
  author={Jiang, Zhengyao and Schmidt, Dominik and Srikanth, Dhruv and Xu, Dixing and Kaplan, Ian and Jacenko, Deniss and Wu, Yuxiang},
  journal={arXiv preprint arXiv:2502.13138},
  year={2025}
}

@inproceedings{agrawal2026gepa,
  title={{GEPA}: Reflective prompt evolution can outperform reinforcement learning},
  author={Agrawal, Lakshya A and Tan, Shangyin and Soylu, Dilara and Ziems, Noah and Khare, Rishi and Opsahl-Ong, Krista and Singhvi, Arnav and Shandilya, Herumb and Ryan, Michael J and Jiang, Meng and Potts, Christopher and Sen, Koushik and Dimakis, Alexandros G and Stoica, Ion and Klein, Dan and Zaharia, Matei and Khattab, Omar},
  booktitle={International Conference on Learning Representations},
  pages={8479--8565},
  year={2026}
}
\bibliographystyle{colm2026_conference}

\appendix
\section{Supplementary Material}

\ifcolmsubmission
We provide code and the best kernels generated by \METHOD{} for the KernelBench Level 2 and Level 3 tasks in the Supplementary Material.
\else
We open-source the implementation of \METHOD{} in \url{https://github.com/StigLidu/AdaExplore}.
\fi

\section{Evaluation on FlashInfer-Bench}
\label{app:flashinfer_bench}

\subsection{Setup}

KernelBench evaluates kernel runtime optimization on general-purpose PyTorch operator rewrites. To further assess performance on real-world LLM serving workloads, we evaluate \METHOD{} on FlashInfer-Bench~\citep{xing2026flashinfer}. Its kernel tasks are extracted from production inference pipelines with input shapes captured from deployed models (e.g., Llama-3.1-8B, Qwen3-30B-A3B), and expert-written FlashInfer CUDA kernels, the same implementations used in production serving frameworks such as SGLang, serve as performance baselines.

We select three kernel definitions spanning different operation types and optimization difficulty:
\begin{itemize}[nosep,leftmargin=*]
    \item \textbf{Fused Add RMSNorm} (\texttt{fused\_add\_rmsnorm\_h2048}): A memory-bound element-wise kernel that computes residual addition followed by RMS normalization.
    \item \textbf{GEMM} (\texttt{gemm\_n128\_k2048}): A general matrix multiplication $C = A B^\top$ with $N{=}128$, $K{=}2048$, captured from the MoE gate of Qwen3-30B-A3B.
    \item \textbf{GQA Paged Decode} (\texttt{gqa\_paged\_decode\_h32\_kv8\_d128\_ps1}): A grouped-query attention decode kernel with paged KV cache, captured from Llama-3.1-8B.
\end{itemize}
We run \METHOD{} with the same MCTS configuration as in the main experiments (50 steps, GPT-5-mini) on an NVIDIA B200 GPU, corresponding to a relatively small test-time compute budget.
For baselines, we benchmark against the PyTorch eager reference and, where available, the FlashInfer CUDA kernel or cuBLAS (\texttt{torch.mm}).

\subsection{Results}

\begin{table}[h]
\centering
\small
\setlength{\tabcolsep}{4pt}
\begin{tabular}{lcccccc}
\toprule
\textbf{Kernel} & \textbf{Steps} & \textbf{Correct} & \textbf{Compile Err.} & \textbf{Best Time} & \textbf{vs PyTorch} & \textbf{vs Expert} \\
\midrule
\texttt{fused\_add\_rmsnorm} & 50 & 16 (32\%) & 34 (68\%) & 0.0075\,ms & \textbf{7.22$\times$} & \textbf{1.75$\times$} \\
\texttt{gemm\_n128\_k2048} & 50 & 23 (46\%) & 25 (50\%) & 0.028\,ms & 0.42$\times$ & 0.42$\times$ \\
\texttt{gqa\_paged\_decode} & 50 & 11 (22\%) & 34 (68\%) & 0.682\,ms & \textbf{18.17$\times$} & 0.15$\times$ \\
\bottomrule
\end{tabular}
\caption{\textbf{FlashInfer-Bench Results} (50-step MCTS, GPT-5-mini, NVIDIA B200). ``vs Expert'' reports the ratio of the expert baseline time (FlashInfer or cuBLAS) to the best \METHOD{}-generated kernel time; values $>1$ indicate the generated kernel is faster.}
\label{tab:flashinfer_bench}
\end{table}

Table~\ref{tab:flashinfer_bench} summarizes the results. We highlight several findings:

\paragraph{RMSNorm: Surpassing Expert-written CUDA.}
The best generated kernel achieves \textbf{1.75$\times$} speedup over the FlashInfer CUDA implementation and 7.22$\times$ over the PyTorch reference.
The generated kernel (shown in Figure~\ref{fig:rmsnorm_kernel}) loads the entire hidden dimension ($H{=}2048$) in a single tile (\texttt{BLOCK=2048}), performing the fused add, variance reduction, and scaling entirely in registers.
This demonstrates that LLM agents can discover hardware-efficient implementations that exceed expert-tuned CUDA for memory-bound workloads.

\paragraph{GEMM: Fundamentally Hard Due to Heavy Human Optimization.}
The best genuine Triton GEMM kernel achieves 0.42$\times$ of cuBLAS performance.
This result is not surprising, as GEMM kernels have been extensively optimized over decades, incorporating sophisticated tiling strategies, scheduling, and hardware-specific instructions in vendor libraries such as cuBLAS.
The difficulty here is therefore intrinsic to the problem setting: improving over such heavily engineered baselines remains fundamentally challenging.

\begin{figure}[h]
\centering
\begin{lstlisting}[style=prompt]
@triton.jit
def _fused_add_rmsnorm_kernel(
    x_ptr, r_ptr, w_ptr, out_ptr,
    stride_xm, stride_xn, stride_rm, stride_rn,
    stride_w, stride_outm, stride_outn,
    M, N, EPS, BLOCK: tl.constexpr):
    pid = tl.program_id(0)
    if pid >= M:
        return
    offs = tl.arange(0, BLOCK)
    mask = offs < N

    x_val = tl.load(x_ptr + pid * stride_xm + offs * stride_xn, mask=mask, other=0.0)
    r_val = tl.load(r_ptr + pid * stride_rm + offs * stride_rn, mask=mask, other=0.0)
    x_f = tl.cast(x_val, tl.float32) + tl.cast(r_val, tl.float32)

    sumsq = tl.sum(x_f * x_f, 0)
    inv_rms = 1.0 / tl.sqrt(sumsq / tl.cast(N, tl.float32) + tl.cast(EPS, tl.float32))

    w_val = tl.load(w_ptr + offs * stride_w, mask=mask, other=0.0)
    y = x_f * tl.cast(w_val, tl.float32) * inv_rms
    tl.store(out_ptr + pid * stride_outm + offs * stride_outn,
             tl.cast(y, tl.bfloat16), mask=mask)
\end{lstlisting}
\caption{Best \METHOD{}-generated fused add RMSNorm kernel (1.75$\times$ faster than FlashInfer). The kernel processes one row per program, loading the entire hidden dimension in a single tile of \texttt{BLOCK=2048} elements and performing the reduction in registers.}
\label{fig:rmsnorm_kernel}
\end{figure}

\paragraph{GQA: large gains over PyTorch, but still far from expert performance.}
While only 11 out of 50 generated kernels pass correctness checks, the best kernel achieves a substantial \textbf{18.17$\times$ speedup over PyTorch}.
However, it remains \textbf{6.5$\times$ slower} than the FlashInfer expert implementation.

The generated kernels use a two-pass softmax that sequentially scans all KV tokens twice per head.
In contrast, the FlashInfer baseline on the B200 employs a warp-specialized XQA kernel with 4 cooperating CTAs, single-pass online softmax, TMA-accelerated paged KV cache access, FP8 attention weight quantization, and Blackwell QMMA instructions.

These advanced mechanisms on B200 highlight the importance of incorporating additional knowledge and reference skills, which we leave as future work for agent-based kernel runtime optimization.

\subsection{Case Study: Best RMSNorm Kernel}
Figure~\ref{fig:rmsnorm_kernel} presents the best kernel found by \METHOD{}.

\begin{figure*}[!ht]
\begin{AIbox}{Training Program Synthesis Prompt}
\small
{\color{blue}\bf System:}
{

\textbf{Task Description}

You are tasked with creating a \textbf{new, complex} PyTorch kernel module that combines:

\verb|  |* Patterns and structures from the provided example files

\verb|  |* PyTorch layers and operations from the provided layer list

Your goal is to create a more complex module that:

\verb|  |* Follows the same code structure as the examples (\verb|Model| class, \verb|get_inputs|, \verb|get_init_inputs|, etc.)

\verb|  |* Incorporates one or more of the provided PyTorch layers in creative ways

\verb|  |* Creates a functionally different computation pattern with a similar difficulty level as the examples above (3--6 operations are recommended)

\verb|  |* Maintains proper PyTorch patterns and code quality

Code structure requirements:

\verb|  |* A \verb|Model| class inheriting from \verb|nn.Module|

\verb|  |* An \verb|__init__| method if needed for parameters

\verb|  |* A \verb|forward| method with the computation

\verb|  |* A \verb|get_inputs()| function

\verb|  |* A \verb|get_init_inputs()| function

\verb|  |* Configuration variables

\textbf{Example Files (for structure reference)}

[Example 1]

[Example 2]

[Example 3]

\textbf{Available PyTorch Layers/Operations to Incorporate}

[Operation 1]

[Operation 2]

[Operation 3]

Generate a \textbf{new, complex} example that:

\verb|  |* Uses the structure from the examples above

\verb|  |* Incorporates one or more of the provided PyTorch layers

\verb|  |* Creates a more complex computation pattern (e.g., combining multiple layers, using different tensor shapes, etc.)

\verb|  |* Is functionally distinct from the provided examples

\verb|  |* Maintains a similar or higher complexity level as the examples

Return \textbf{only} the Python code, wrapped in \verb|```python| code blocks. Do not include any explanations or markdown formatting outside the code block.

}
\end{AIbox} 
\caption{\textbf{Training Program Synthesis Prompt.}}
\label{fig:synthetic_prompt}
\end{figure*}
\section{Detailed Description of Synthesized Training Tasks}

\subsection{Dataset Synthesis}
\label{app:dataset_synthesis}
To expand the dataset, we use mutation-based prompting in the spirit of Evol-Instruct \citep{xu2025wizardlmempoweringlargepretrained}. In each iteration, we sample three seed task examples and operators from the PyTorch documentation~\footnote{\url{https://docs.pytorch.org/docs/stable/nn.html}}, and prompt GPT-5 to generate a new PyTorch module by mutating and recombining existing patterns and operators. We execute the generated code on the synthesized test tensors and discard any samples that cause errors. This mutation process introduces finer-grained, low-level variation and yields samples that remain closer in complexity to our seed task examples. The prompt can be found in Figure~\ref{fig:synthetic_prompt}.


Our agent uses these training tasks to collect cross-task skill memory from attempts to implement kernels. For this purpose, we found that mutation-based synthetic data provided a richer signal, likely because it involved more complex use of low-level operations rather than common higher-level building blocks. This forced the agent into edge cases, where it was more likely to make errors that could be stored in memory.


\subsection{Examples of Training Tasks}
\label{app:syn_data_example}
\begin{table}[!ht]
\scriptsize
\centering
\begin{tabular}{p{0.96\linewidth}}
\toprule
\textbf{Examples of Synthesized Task Description 1} \\
\midrule
\begin{minipage}[t]{\linewidth}
\begin{verbatim}
    Complex 2D feature-transform block combining pointwise reduction, depthwise convolution,
    batch normalization, Hardswish activations, and a tanh-based global gating mechanism.
    The pattern creates a residual-style transformation with an input-conditioned gate.

    Computation graph:
      identity = skip(x)
      x_reduced = conv1x1(x) -> bn -> hardswish
      x_dw = depthwise_conv(x_reduced) -> bn -> hardswish
      x_proj = conv1x1(x_dw) -> bn
      gate = tanh(gate_conv(adaptive_avg_pool(x)))
      out = hardswish( x_proj * gate + identity )
\end{verbatim}
\end{minipage}
\\
\midrule
\textbf{Examples of Synthesized Task Description 2} \\
\midrule
\begin{minipage}[t]{\linewidth}
\begin{verbatim}
    Complex 3D feature gating module.

    This model accepts a 5D tensor (N, C, D, H, W), applies replication padding,
    computes global average and max pooled channel descriptors from the padded tensor,
    combines them into a gating vector which is passed through a learned linear
    projection and a Sigmoid to produce a channel-wise gate. The gate is applied
    to the original (unpadded) input and then a Threshold nonlinearity is used.
    Finally, spatial dimensions are summed to produce an (N, C) output.

    This pattern demonstrates the use of nn.ReplicationPad3d, nn.Sigmoid, and
    nn.Threshold combined with tensor reductions and a small learnable projection.
\end{verbatim}
\end{minipage}
\\
\midrule
\textbf{Examples of Synthesized Task Description 3} \\
\midrule
\begin{minipage}[t]{\linewidth}
\begin{verbatim}
    A composite model that:
    - Applies BatchNorm2d to a 4D input (B, C, H, W)
    - Converts spatial locations (H*W) into a sequence
    - Maps per-location channels into an RNN input space via a Linear layer
    - Runs an nn.RNNCell across the spatial sequence (treating flattened H*W as time steps)
    - Uses Tanhshrink non-linearity between Linear -> RNNCell
    - Projects RNN hidden states back to output channels and reshapes to (B, out_channels, H, W)
\end{verbatim}
\end{minipage}
\\
\bottomrule
\end{tabular}
\caption{\textbf{Examples of Synthesized Training Tasks.}}\label{tab:code-gen}
\end{table}
Table~\ref{tab:code-gen} shows three examples of synthesized training tasks. Starting from real PyTorch seed programs, we ask the model to generate new tasks that preserve the input--output interface while composing operators into new computation patterns. The examples illustrate the kind of structural variation we want: one combines pointwise and depthwise convolutions with gating, another builds a 3D channel-gating module with pooling and thresholding, and the third turns spatial features into a recurrent sequence processed by an \texttt{RNNCell}. Together, these examples show that synthesis produces tasks that remain realistic in structure while exposing the agent to more diverse operator compositions and implementation challenges. 


\paragraph{Cross-task Skill Memory Statistics.}
\label{app:memory_analysis}
We synthesize 200 training tasks (examples in Appendix~\ref{app:syn_data_example}) and run \METHOD{} to collect error experiences in 25 steps per task. Table~\ref{tab:cross_problem_memory} shows that 1,178 raw experiences are observed during self-exploration with GPT-5-mini as the base model, but only 174 remain after semantic deduplication, and only 78 are retained as transferable high-frequency hints after filtering (examples in Table~\ref{tab:memory}). This high redundancy suggests that failures in low-level kernel generation are concentrated around a relatively small set of recurring constraints.

The retained memory spans multiple categories, with kernel syntax and DSL constraints accounting for the largest portion, followed by Python/environment issues and memory/indexing errors. This compact final memory filters out task-specific noise while preserving stable, reusable heuristics.

\subsection{Examples of Cross-task Skill Memory}
Tables~\ref{tab:memory} and~\ref{tab:memory_lowfreq} show the highest- and lowest-frequency entries in the learned cross-task skill memory. These entries summarize recurring failure patterns observed during experience collection and convert them into short, actionable reminders. A clear pattern emerges: high-frequency items tend to capture stable and broadly useful constraints. For example, frequent rules such as treating \texttt{tl.float32} as a function, using unsupported Triton indexing patterns, or omitting required constexpr launch parameters correspond to recurring API and language constraints that arise across many tasks. In contrast, low-frequency items are more often tied to narrow implementation details, one-off bugs, or noisy summaries of rare failures, such as attempting to connect to \texttt{0.0.0.0:12017}, computing fan-in/fan-out for a tensor with fewer than two dimensions, or compiling a very specific statement involving \texttt{tl.zeros((patch\_dim\_out,), dtype=tl.float32)}. This contrast supports our filtering strategy: high-frequency persistent-skill-memory entries are more likely to be transferable across tasks, whereas very low-frequency entries are more likely to contain noise or even incorrect guidance.
\begin{table}[t]
    \centering
    \scriptsize
    \setlength{\tabcolsep}{4pt}
    \begin{tabular}{p{0.04\linewidth} p{0.08\linewidth} p{0.78\linewidth}}
    \toprule
    \# & Freq. & Skill Description \\
    \midrule
    1 & 138 & You cannot call \texttt{tl.float32} as a function inside a Triton kernel. \\
    2 & 79 & You cannot have an \texttt{else:} statement that produces invalid Python syntax in the generated module. \\
    3 & 72 & You cannot pass a \texttt{torch.cuda.FloatTensor} as the indices argument to \texttt{torch.embedding\_bag}. \\
    4 & 61 & You cannot index a Triton tensor with a constexpr index inside a Triton kernel at compile time. \\
    5 & 36 & You cannot access \texttt{triton.language.math.tanh}. \\
    6 & 36 & You cannot index a Triton tensor with an \texttt{int32[]} tensor index inside a Triton kernel during compilation. \\
    7 & 34 & You cannot reference the name \texttt{PADDING\_LEFT} when it is not defined. \\
    8 & 31 & You cannot call \texttt{tl.load(x\_ptr + x\_addrs, mask=mask\_x, other=0.0)} inside the Triton kernel shown. \\
    9 & 31 & You cannot call the Triton kernel without passing the required constexpr parameter \texttt{BLOCK}. \\
    10 & 31 & You cannot define class \texttt{ModelNew} that inherits from \texttt{Model} when \texttt{Model} is not defined. \\
    11 & 27 & You cannot reference the name \texttt{Model} inside \texttt{ModelNew.\_\_init\_\_}. \\
    12 & 25 & You cannot index \texttt{tl.zeros((1,), dtype=tl.float32)} with \texttt{[0]} inside a Triton \texttt{@jit} kernel. \\
    13 & 25 & You cannot pass an indices tensor with float elements to \texttt{torch.nn.functional.max\_unpool3d}. \\
    14 & 24 & You cannot use \texttt{tl.load(inp\_ptr + in\_index)} in this Triton kernel. \\
    15 & 22 & You cannot call \texttt{float()} on a tensor such as \texttt{kernel\_vol} inside a Triton kernel. \\
    16 & 20 & You cannot use Triton builder operations that require a \texttt{\_builder} argument outside functions compiled with \texttt{@triton.jit}. \\
    17 & 19 & You cannot pass an integer as the grid parameter when launching a Triton kernel. \\
    18 & 18 & You cannot use the \texttt{break} statement inside a Triton kernel. \\
    19 & 16 & You cannot call \texttt{.reshape(-1)} inside a Triton kernel. \\
    20 & 14 & You cannot call \texttt{tl.arange(0, BLOCK)} inside the Triton kernel \texttt{\_mean\_invstd\_kernel}. \\
    \bottomrule
    \end{tabular}
    \caption{\textbf{Top-20 Cross-Task Skill Memory Entries Ranked by Frequency.}}\label{tab:memory}
\end{table}

\begin{table}[t]
    \centering
    \scriptsize
    \setlength{\tabcolsep}{4pt}
    \begin{tabular}{p{0.04\linewidth} p{0.08\linewidth} p{0.78\linewidth}}
    \toprule
    \# & Freq. & Skill Description \\
    \midrule
    1 & 1 & You cannot use a keyword assignment such as \texttt{num\_warps=8} inside the square-bracket kernel launch configuration when calling a Triton kernel. \\
    2 & 1 & You cannot call \texttt{TritonPointwise.run} with a weight tensor \texttt{W} that has 0 dimensions. \\
    3 & 1 & You cannot compute fan-in and fan-out for a tensor with fewer than 2 dimensions. \\
    4 & 1 & You cannot establish an HTTP connection to \texttt{0.0.0.0:12017} for normal evaluation. \\
    5 & 1 & You cannot access the shape of an uninitialized parameter or buffer. \\
    6 & 1 & You cannot use the \texttt{**} operator to raise a Triton tensor inside a Triton kernel. \\
    7 & 1 & You cannot call \texttt{tl.store} with a 1-element tensor (shape \texttt{[1]}) as the value in this Triton kernel. \\
    8 & 1 & You cannot call \texttt{ModelNew.\_\_init\_\_} with the keyword argument \texttt{mid\_channels}. \\
    9 & 1 & You cannot call \texttt{apply()} with keyword arguments. \\
    10 & 1 & You cannot perform simultaneous multiple assignment. \\
    11 & 1 & You cannot access \texttt{triton.language.concat}. \\
    12 & 1 & You cannot pass a \texttt{num\_channels} keyword argument to \texttt{torch.nn.functional.group\_norm}. \\
    13 & 1 & You cannot pass an \texttt{int} as the \texttt{weight} argument to \texttt{torch.group\_norm}. \\
    14 & 1 & You cannot call \texttt{torch.add} with an \texttt{out=} argument when one of the arguments requires gradients. \\
    15 & 1 & You cannot unpack \texttt{rms\_view.stride()} into two variables. \\
    16 & 1 & You cannot call \texttt{torch.matmul} with an \texttt{out=} argument when one of the arguments requires gradients. \\
    17 & 1 & You cannot convert a tensor with more than one element to a Python scalar using \texttt{int(...)}. \\
    18 & 1 & You cannot access the attribute \texttt{mod} on the \texttt{triton.language} module. \\
    19 & 1 & You cannot call \texttt{tl.num\_programs()} inside a Triton kernel at compile time. \\
    20 & 1 & You cannot compile a Triton kernel that contains the statement \texttt{accs = tl.zeros((patch\_dim\_out,), dtype=tl.float32)}. \\
    \bottomrule
    \end{tabular}
    \caption{\textbf{Bottom-20 Cross-Task Skill Memory Entries Ranked by Frequency.}}\label{tab:memory_lowfreq}
\end{table}

\subsection{Static Documentation}
\label{app:static_doc}
To test whether the cross-skill memory can be replaced by a Triton reference that could be written once from the documentation, we construct a static-documentation baseline and substitute it for the skill memory.

The static documentation baseline is constructed from the public API of the Triton language by introspection from the same Triton installation (Triton 3.1.0) that compiles the generated kernels. It contains 122 symbols from \verb|triton.language| and 20 symbols from \verb|triton.language.math|, each with its call signature. It is comparable in size to the skill memory it replaces (1,616 versus 1,655 cl100k tokens) and is presented as an API reference rather than as observed failures.

\subsection{Generalizability of Cross-Task Skill Memory}
To further evaluate the generalizability of \METHOD{}, we conduct experiments on TritonBench-T~\citep{li2025tritonbench}. This benchmark requires models to generate behaviorally equivalent, interface-compatible Python/Triton implementations from function semantic descriptions and interface specifications. It consists of 166 tasks covering basic element-wise operators, reductions, linear algebra, and various fused operators. We evaluate single-pass generation with and without cross-task skill memory collected from synthetic KernelBench questions, as well as 5-step and 10-step \METHOD{} on TritonBench-T. The results are shown in Table~\ref{tab:tbt}. Notably, although the cross-task skill memory is collected from synthetic KernelBench questions, it generalizes well to TritonBench-T. With this cross-task skill memory, GPT-5-mini improves its single-pass generation accuracy from 54\% to 82\%. \METHOD{} further boosts performance, with the 10-step variant reaching 97\% accuracy and 24\% Fast@1.2.
\label{app:tritonbench}
\begin{table}[!t]
    \centering
    \small
    \setlength{\tabcolsep}{6pt}
    \renewcommand{\arraystretch}{1.1}
    \begin{tabular}{lccc}
    \toprule
    \textbf{TritonBench-T} & \textbf{Acc.} & \textbf{Fast@1.2} \\
    \midrule
    \textbf{GPT-5-mini} & 54\% & 7\% \\
    \textbf{GPT-5-mini w. SM} & 82\% & 7\% \\
    \textbf{AE + GPT-5-mini (5 Steps)} & \underline{95\%} & \underline{19\%} \\
    \textbf{AE + GPT-5-mini (10 Steps)} & \textbf{97\%} & \textbf{24\%} \\
    \bottomrule
    \end{tabular}
    \caption{\textbf{TritonBench-T Performance.} The best result in each column is \textbf{bold}, and the second best is \underline{underlined}.}
    \label{tab:tbt}
    \end{table}

\subsection{Full Pipeline on a Small Open-Weight Model}
\label{app:qwen_pipeline}
To test whether \METHOD{} generalizes beyond proprietary models, we run the full pipeline on Qwen3-Coder-Next~\citep{cao2026qwen3}, an open-weight MoE with only 3B active parameters (Table~\ref{tab:qwen_pipeline}). Cross-task skill memory alone raises correctness from 7\% to 91\% and speedup from 0.09$\times$ to 1.54$\times$, and the Explore stage further improves them to 96\% and 1.57$\times$. This indicates that \METHOD{} is not tied to large proprietary models.
\begin{table}[h]
\centering
\small
\setlength{\tabcolsep}{6pt}
\renewcommand{\arraystretch}{1.1}
\begin{tabular}{lcccc}
\toprule
\textbf{Method (Qwen3-Coder-Next)} & \textbf{Acc.} & \textbf{Speedup} & \textbf{Fast@1.2} & \textbf{Fast@2} \\
\midrule
Single Pass & 7\% & 0.09$\times$ & 2\% & 1\% \\
PS (50 Steps) & 72\% & 1.12$\times$ & 25\% & 10\% \\
PS \textit{w. SM} (50 Steps) & 91\% & 1.54$\times$ & 24\% & 15\% \\
\textbf{\METHOD{} (50 Steps)} & \textbf{96\%} & \textbf{1.57$\times$} & \textbf{33\%} & \textbf{18\%} \\
\bottomrule
\end{tabular}
\caption{\textbf{Full \METHOD{} Pipeline on Qwen3-Coder-Next} (KernelBench Level-2, 50 steps). Qwen3-Coder-Next is an open-weight MoE with only 3B active parameters. \textit{w. SM} denotes augmenting with our cross-task skill memory.}
\label{tab:qwen_pipeline}
\end{table}

\subsection{Cross-Model Memory Transfer}
\label{app:cross_model}
Table~\ref{tab:knowledge_base_performance} uses a separate memory for each model. We further test whether a memory built by one model transfers to another by applying GPT-5-mini's memory to GPT-5 and DeepSeek-R1~\citep{guo2025deepseekr1} on KernelBench L2 (Table~\ref{tab:cross_model_transfer}). Transfer is not universal: GPT-5 benefits substantially (Pass@1 $51\%\rightarrow65\%$), whereas DeepSeek-R1 barely changes ($33\%\rightarrow34\%$), suggesting that cross-model transfer depends on the similarity between the source and target models.
\begin{table}[h]
\centering
\small
\setlength{\tabcolsep}{8pt}
\renewcommand{\arraystretch}{1.1}
\begin{tabular}{lcc}
\toprule
\textbf{Pass@1 on KernelBench L2} & \textbf{GPT-5} & \textbf{DeepSeek-R1} \\
\midrule
Without memory & 51\% & 33\% \\
With GPT-5-mini memory & \textbf{65\%} & 34\% \\
\bottomrule
\end{tabular}
\caption{\textbf{Cross-Model Memory Transfer.} We apply the cross-task skill memory constructed by GPT-5-mini to GPT-5 and DeepSeek-R1~\citep{guo2025deepseekr1}. GPT-5 benefits substantially, whereas DeepSeek-R1 shows almost no gain, suggesting that transfer effectiveness depends on the similarity between the source and target models.}
\label{tab:cross_model_transfer}
\end{table}

\section{Method Details}
\label{app:method_details}
\begin{algorithm}[th]
\caption{Learning Skills from Failures (Adapt)}
\label{alg:adapt}\small
\begin{algorithmic}[1]
\STATE {\bfseries Input:} synthesized task set $\mathcal{D}_{\text{syn}}$; coding agent $\pi$; frequency threshold $O$
\STATE {\bfseries Output:} filtered cross-task skill memory $\mathcal{M}$
\STATE Initialize raw memory $\widetilde{\mathcal{M}} \leftarrow \emptyset$
\FOR{each synthesized task $x \in \mathcal{D}_{\text{syn}}$}
    \STATE Generate kernel candidates $\{y_1, ..., y_k\} \sim \pi(\cdot \mid x, \widetilde{\mathcal{M}})$
    \STATE Execute $\{y_1, ..., y_k\}$ against the reference implementation of $x$ and collect feedback $\{f_1, ..., f_k\}$
    \FOR{each candidate $y_i$}
    \IF{$y_i$ fails compilation or execution}
        \STATE Summarize $(y_i, f_i)$ into a concise constraint rule $m$
        \IF{$m$ is semantically matched to an existing memory item $m' \in \widetilde{\mathcal{M}}$}
            \STATE Increase the frequency count of $m'$
        \ELSE
            \STATE Add $m$ to $\widetilde{\mathcal{M}}$ with frequency count $1$
        \ENDIF
    \ENDIF
\ENDFOR
\ENDFOR
\STATE Filter $\mathcal{M} \leftarrow \{m \in \widetilde{\mathcal{M}} \mid \mathrm{freq}(m) \ge O\}$
\RETURN $\mathcal{M}$
\end{algorithmic}
\end{algorithm}
\begin{algorithm}[th]
\caption{Diversity-Preserving Tree Search (Explore)}
\label{alg:evolve}\small
\begin{algorithmic}[1]
\STATE {\bfseries Input:} target problem $x$; coding agent $\pi$; cross-task skill memory $\mathcal{M}$; search budget $T$; recent-window size $C_{\text{recent}}$; pool size $C_{\text{pool}}$; large-step probability $p_{\text{large}}$
\STATE {\bfseries Output:} best kernel found $y^\star$
\STATE Initialize search tree $\mathcal{T}$ with a virtual root node $s_{\mathrm{root}}$
\FOR{$t=1$ to $T$}
    \STATE Select a node $s$ from $\mathcal{T}$ using the UCT-style rule in Section~\ref{sec:context_mem}
    \STATE Construct working memory $W_s$ from the most recent $C_{\text{recent}}$ states on the path to $s$
    \STATE Build a set of representative kernels $\mathcal{K}_{\text{pool}}$ by keeping at most one kernel from each connected segment of consecutive small steps on the current path
    \STATE Sample at most $C_{\text{pool}}$ representative kernels from $\mathcal{K}_{\text{pool}}$.
    \IF{$s = s_{\mathrm{root}}$}
        \STATE Set $a \leftarrow \textsc{LargeStep}$
    \ELSE
        \STATE Set $a=\textsc{LargeStep}$ with probability $p_{\text{large}}$ and $a=\textsc{SmallStep}$ otherwise
    \ENDIF
    \IF{$a=\textsc{LargeStep}$}
        \STATE Clear $W_s$
    \ELSE
        \STATE Clear $\mathcal{K}_{\text{pool}}$
    \ENDIF
    \STATE Generate a new kernel candidate $y' \sim \pi(\cdot \mid x, \mathcal{M}, W_s, \mathcal{K}_{\text{pool}}, a)$
    \STATE Execute $y'$ and obtain correctness and performance feedback
    \STATE Add a new child node $s'$ under $s$ with edge type $a$
    \STATE Store $y'$, feedback, and score in $s'$
    \STATE Update visit counts and node statistics on the path from $s'$ to the root
\ENDFOR
\STATE Return the highest-performing correct kernel in $\mathcal{T}$ as $y^\star$
\end{algorithmic}
\end{algorithm}
\subsection{Algorithm Pseudocode}
The pseudocode of \textsc{Adapt} is shown in Algorithm~\ref{alg:adapt}.
The pseudocode of \textsc{Explore} is shown in Algorithm~\ref{alg:evolve}.
\subsection{Action Description}
\label{app:actions}
\begin{figure*}[!ht]
\begin{AIbox}{Shared Search Context}
\small
{\color{blue}\bf System:}
{

\textbf{Problem Statement}

You write custom kernels to replace the PyTorch operators in the given architecture to get speedups.

You have complete freedom to choose which operators to replace. You may replace some operators with custom kernels and leave others unchanged. You may replace multiple operators, consider operator fusion opportunities (e.g., combining \verb|matmul| and \verb|relu|), or introduce algorithmic changes such as online softmax. You may also reorder mathematically equivalent operations to enable better fusion or memory access patterns.

\textbf{Hardware Information}

\begin{lstlisting}[style=prompt]
[Hardware Information]
\end{lstlisting}

\textbf{Example Formats}

Here is an example showing the syntax of inline custom Triton kernels in PyTorch.

\textbf{Example Input}

\begin{lstlisting}[style=prompt]
[example task input]
\end{lstlisting}

\textbf{Example Output with Custom Triton Kernels}

\begin{lstlisting}[style=prompt]
[example task output with custom Triton kernels]
\end{lstlisting}

\textbf{Experience Guidance}

Here is some experience guidance that you should keep in mind:

{\color{blue}\textless cross-task skill memory \textgreater}

\textbf{Task Description}

\begin{lstlisting}[style=prompt]
[Task Description]
\end{lstlisting}

\textbf{Kernel Pool}

Here are some community-developed kernels and their runtime metrics. These represent strong baseline implementations for the given task.

{\color{blue}\textless representative kernels and metrics \textgreater}

\textbf{Previous Kernels and Metrics}

Previously, you have generated the following custom kernels and got the following runtime metrics:

{\color{blue}\textless working memory \textgreater}

}
\end{AIbox}
\caption{\textbf{Shared Search Context Prompt.}}
\label{fig:context_prompt}
\end{figure*}

All actions are performed by the coding agent and share the same context prompt, as shown in Figure~\ref{fig:context_prompt} (some adjustments are applied to ensure readability). If some parts of the information are missing, this part of the prompt will be removed from the context. We use two action types during tree search: \textbf{Large Step} and \textbf{Small Step}, described as follows:
\begin{figure*}[!ht]
\begin{AIbox}{Large-Step Reconstruction Prompt}
\small
{\color{blue}\bf System:}
{

Your goal is to generate a kernel that \textbf{outperforms} all kernels in the pool.

You are allowed to reuse, adapt, or directly build upon any kernel in the pool. You may combine ideas, modify implementations, or start from the best-performing kernel and improve it further.

Objective:

\verb|  |* Minimize runtime as much as possible

\verb|  |* Beat the fastest kernel in the pool

Now generate a kernel that can potentially outperform the best existing kernel and achieves the lowest possible runtime.

}
\end{AIbox}
\caption{\textbf{Large-step Reconstruction Prompt.}}
\label{fig:large_step_prompt}
\end{figure*}

\paragraph{Large Step.} Large step performs \emph{reconstruction}: instead of making a small patch to the current kernel, it prompts the model to generate a new kernel structure conditioned on the shared search context and the large-step objective. The prompt can be found in Figure~\ref{fig:large_step_prompt}. In practice, the agent will regenerate a structurally different kernel from those in the representative pool.

\paragraph{Small Step.}
\begin{figure*}[!ht]
\begin{AIbox}{Small-Step Tuning Prompt}
\small
{\color{blue}\bf System:}
{

\textbf{Goal}

Perform small, localized updates to the last version of the custom kernels using string replacement. Keep the overall model interface unchanged, but you may reorder or restructure internal operations when mathematically equivalent for better performance.

This action proceeds in two stages: first identify concrete modifications based on the provided tuning guidance, and then output one or more larger code patches that implement those modifications.

\textbf{Editing Requirements}

\verb|  |* The \verb|old_str| must match the file content exactly, including whitespace and indentation

\verb|  |* Each \verb|old_str| must uniquely identify one code region

\verb|  |* Each \verb|new_str| must replace the corresponding \verb|old_str| with edited code

\verb|  |* Prefer to output all edits in a single response

\textbf{Output Format}

For each edit, provide:

\verb|  |* \verb|<reasoning_i>|: why this edit is needed

\verb|  |* \verb|<old_str_i>|: the exact old code block

\verb|  |* \verb|<new_str_i>|: the replacement code block

\textbf{Tuning Guidance}

First provide \textbf{review guidance} that explains the intended modifications, and then perform \textbf{code editing} by outputting the edit blocks.

}
\end{AIbox}
\caption{\textbf{Small-step Tuning Prompt.}}
\label{fig:small_step_prompt}
\end{figure*}

The small step performs local tuning on the current kernel. Instead of regenerating a new structure, it first identifies concrete modifications or improvement plans by providing guidance, and then applies one or more code patches to improve correctness or runtime performance. Each patch is specified as an \texttt{old\_str}/\texttt{new\_str} pair: \texttt{old\_str} must exactly match a unique code region in the current kernel, and \texttt{new\_str} provides the corresponding replacement block. The prompt used for this step is shown in Figure~\ref{fig:small_step_prompt}.

The self-generated review guidance, introduced before editing in the small-step process, broadens the scope of potential modifications and encourages more coherent updates. However, we observe that such guidance can sometimes reduce the correctness rate of the generated kernels, as it often introduces complex or abstract instructions that the model has difficulty reliably following. To balance this trade-off, we conditionally enable review guidance based on the current state of the working memory. Specifically, when all kernels in the working memory exhibit correctness issues, we disable review guidance and instead focus the model on fixing errors using execution feedback. In contrast, when at least one correct kernel is present, we enable review guidance to encourage higher-quality and more globally consistent edits.

\section{Experiment Details}
\begin{table}[th]
\centering
\small
\setlength{\tabcolsep}{4pt}
\begin{tabular}{ccc|cccccc}
\toprule
\multicolumn{3}{c|}{\textsc{Adapt}} & \multicolumn{6}{c}{\textsc{Explore}} \\
\midrule
\# tasks & \# candidates & $O$ &  $C_{\text{recent}}$ & $C_{\text{pool}}$ & $c_{\mathrm{explore}}$ & $c_{\mathrm{expand}}$ & $p_{\text{large}}$ & $T_{\text{LLM}}$ \\
\midrule
200 & 25 & 3 & 5 & 5 & 0.3 & 0.3/0.15 & 0.2 & 1\\
\bottomrule
\end{tabular}
\caption{\textbf{Main Hyperparameter Settings Used in \textsc{Adapt} and \textsc{Explore}.}}
\label{tab:hyperparameters}
\end{table}
\subsection{Hyperparameter Settings}
We summarize the main hyperparameter settings used throughout the paper here. For \textsc{Adapt}, the key choices include the number of synthesized training tasks and the number of kernel candidates sampled per synthesized task, and the frequency threshold $O$ used to retain high-frequency cross-task skill-memory entries. For \textsc{Explore}, the main search hyperparameters include the test-time search budget $T$, the recent-context window size $C_{\text{recent}}$, the maximum number of representative kernels sampled into context $C_{\text{pool}}$, the exploration coefficient $c_{\mathrm{explore}}$ and $c_{\mathrm{expand}}$ in the UCT-style node selection rule, large-step probability $p_{\text{large}}$, and the inference temperature $T_{\text{LLM}}$. Specifically, we observe that harder tasks may require more budget focused on the current branch. Consequently, we set $c_{\mathrm{expand}} = 0.3$ for the KernelBench Level 2 task and $c_{\mathrm{expand}} = 0.15$ for the Level 3 task.

\paragraph{OpenEvolve.} We set the population size to 5, meaning that the evolutionary search maintains five candidate solutions at a time. The migration interval is set to 10, so migration is performed every 10 iterations to promote information sharing across search trajectories.
\label{app:openevolve_hyper}

\paragraph{Test Time.}
The main experiments are conducted from January 2026 to April 2026.

\subsection{Evaluation}
We evaluate each generated kernel against the original PyTorch reference using a unified pipeline.
Given a reference implementation and a candidate optimized version, we instantiate both with identical initial inputs, compare their outputs on the same randomly generated test inputs, and measure their speed.

\paragraph{Remote evaluation server.}
For scalability and robustness, we support remote evaluation via a GPU-backed service.
The server maintains a queue of requests, assigns each to an available GPU, executes the candidate in isolation, and returns structured results, including compilation status, correctness status, runtime statistics, and diagnostic metadata.

\paragraph{Correctness checking.}
We assess correctness by comparing candidate outputs with the reference on identical inputs.
For each evaluation, we fix a base seed and deterministically derive per-trial seeds.
A candidate is considered correct only if it matches the reference in all trials.
For each trial, we first verify that the output shapes match and then compare the values using \texttt{torch.allclose} with absolute and relative tolerances set to $1 \times 10^{-2}$, which aligns with the common settings in DR. Kernel~\citep{liu2026dr} and KernelSkill~\citep{sun2026kernelskill}.
If any trial fails, we record diagnostic statistics (e.g., maximum and mean absolute differences) and mark the candidate as incorrect.

\paragraph{Performance measurement.}
We measure runtime only for candidates that pass correctness checks.
Before timing, we clear the CUDA cache and synchronize the device to minimize interference.
Each kernel is executed with 10 warm-up iterations followed by 100 timed trials.
We report summary statistics (mean, standard deviation, minimum, and maximum runtime).
To reduce noise, we apply symmetric outlier trimming by discarding the fastest and slowest 5\% of runs and computing statistics over the remaining 90 trials.




\subsection{Cost Analysis}
\label{app:cost}

We price every stage uniformly at 16K input and 4,096 output tokens per step, which in our observation is a generous estimate for \METHOD{}: steps average 8.7K input tokens, and refer to Appendix~\ref{app:kernel_growth} for how context length evolves with the budget. At OpenRouter list rates, a step costs \$0.0122 for GPT-5-mini (\$0.25/\$2.00 per million input/output tokens) and \$0.0610 for GPT-5 (\$1.25/\$10.00). The resulting price agrees with billed usage to within 10\% (\$29.28 estimated against \$32.53 measured over a 2,400-step run).

\begin{itemize}
  \item \textbf{Task synthesis} (GPT-5, one-time). 200 accepted tasks, ${\sim}300$ generations including rejected samples, each priced as one step: \textbf{\$18}.
  \item \textbf{Adaptation} (GPT-5-mini, one-time). 200 synthesized tasks $\times$ 25 steps, each charged three times---generating a candidate, summarizing the failure into a skill entry, and deduplicating it against the memory---for 15,000 steps: \textbf{\$183}. The latter two do not fire at every step, so this is an upper bound.
  \item \textbf{Search} (GPT-5-mini, per configuration). Level 2 costs \$61 at $T{=}50$, \$122 at $T{=}100$ and \$244 at $T{=}200$; Level 3 costs \$30 at $T{=}50$ and \$61 at $T{=}100$.
  \item \textbf{Wall-clock.} Dominated by kernel evaluation rather than inference: 100 tasks at 25 steps take 4--5 hours on two A6000s with one kernel timed per GPU at a time.
\end{itemize}

The two one-time stages total \$201 and are shared rather than specific to \METHOD{}: the IR \textit{w. SM} and PS \textit{w. SM} baselines consume the same skill memory and carry the same cost. The comparison that matters is therefore at equal total spend, and \METHOD{} reaches a given speedup for less: \textbf{\$323 at $T{=}100$ against \$445 for IR \textit{w. SM} at $T{=}200$}, at the same speedup.

\subsection{An Observation: Refinement Patches, Search Rewrites}
\label{app:kernel_growth}

While collecting the cost figures below, we noticed a difference in the context length of AdaExplore and iterative refinement. Both cap the retained history at five rounds, yet the prompt of iterative refinement grows with the test-time budget, while ours does not: averaged over Level-2 tasks, iterative refinement's prompt rises from 10.3K input tokens over the first ten steps to 36.6K over steps 51--200, whereas a full \METHOD{} step stays near 10K throughout.

\begin{figure}[h]
    \centering
    \includegraphics[width=0.55\linewidth]{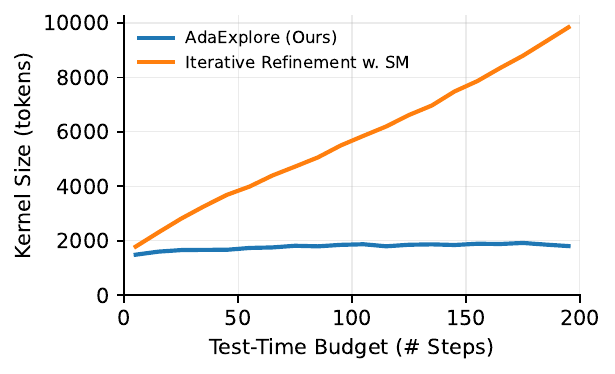}
    \caption{\textbf{Kernel size against test-time budget} on KernelBench Level-2, in cl100k tokens of the emitted source, binned every ten steps and averaged over all tasks. Iterative refinement accretes code along a single chain, while \METHOD{} periodically regenerates from scratch and stays flat.}
    \label{fig:kernel_growth}
\end{figure}

The cause is not the number of retained rounds but their contents. Figure~\ref{fig:kernel_growth} tracks the size of the emitted kernel against the steps: the kernels that iterative refinement carries in context grow from 1.8K tokens at the beginning to 7.7K over steps 101--200, a $4.3\times$ expansion, while \METHOD{}'s stay essentially flat (1.5K to 1.9K, $1.25\times$) despite starting at a comparable size.

 An explanation for this phenomenon is that refinement along a single chain mostly edits and patches an existing implementation, so code accumulates and is rarely removed. In contrast, a $p_{\text{large}}$ rate of \METHOD{}'s steps regenerates a kernel from scratch and resets its size, and the tree retains earlier candidates eligible for selection.

\section{Disclosure of LLM use}
We used LLMs in three parts of this work. First, LLMs were used to synthesize the synthetic training tasks described in the adaptation stage. Second, LLMs served as the coding agents in our experiments, where they generated and refined kernel implementations under execution feedback. Third, LLMs were used to help refine the paper's writing, including wording and clarity.

\end{document}